\pdfoutput=1

\documentclass[11pt]{article}

\usepackage[final]{acl}

\usepackage{times}
\usepackage{latexsym}

\usepackage[T1]{fontenc}

\usepackage[utf8]{inputenc}

\usepackage{microtype}

\usepackage{inconsolata}

\usepackage{graphicx}

\usepackage{tcolorbox}
\usepackage{tikz}
\usepackage[edges]{forest}

\title{Position: LLMs Can be Good Tutors in English Education}

\author{
    Jingheng Ye$^{1,2}$,
    Shen Wang${^1}$,
    Deqing Zou${^2}$,
    Yibo Yan$^{1,3}$,\\
    \textbf{Kun Wang}$^{1}$,
    \textbf{Hai-Tao Zheng}$^{2,4}$\thanks{Corresponding Author.},
    \textbf{Ruitong Liu}$^{2}$,
    \textbf{Zenglin Xu}$^{5}$,\\
    \textbf{Irwin King}${^6}$,
    \textbf{Philip S. Yu}$^{7}$,
    \textbf{Qingsong Wen}$^{1}$\footnotemark[1]\\
    $^1$Squirrel Ai Learning,
    $^2$Tsinghua University,\\
    $^3$HKUST (Guangzhou),
    $^4$Peng Cheng Laboratory,
    $^5$Fudan University,\\
    $^6$The Chinese University of Hong Kong,
    $^7$University of Illinois at Chicago\\
    \texttt{yejh22@mails.tsinghua.edu.cn}
}

\begin{document}
\maketitle

\begin{abstract}
While recent efforts have begun integrating large language models (LLMs) into English education, they often rely on traditional approaches to learning tasks without fully embracing educational methodologies, thus lacking adaptability to language learning. To address this gap, we argue that \textbf{LLMs have the potential to serve as effective tutors in English Education}. Specifically, LLMs can play three critical roles: (1) as \textit{data enhancers}, improving the creation of learning materials or serving as student simulations; (2) as \textit{task predictors}, serving as learner assessment or optimizing learning pathway; and (3) as \textit{agents}, enabling personalized and inclusive education. We encourage interdisciplinary research to explore these roles, fostering innovation while addressing challenges and risks, ultimately advancing English Education through the thoughtful integration of LLMs.

\end{abstract}

\section{Introduction}\label{sec:introduction}

English Education has long been a cornerstone of global education and a critical component of K-12 curricula, equipping students with the linguistic and cultural competencies necessary for an interconnected world~\cite{alhusaiyan2025systematic,katinskaia2025overview}. However, traditional English teaching methods often fall short in addressing the diverse needs of learners~\cite{hou2020foreign}. Challenges such as limited personalization, scalability constraints, and the lack of real-time feedback are particularly pronounced in large classroom settings~\cite{ehrenberg2001class}. Addressing these shortcomings requires innovative approaches that not only enhance the quality of instruction but also adapt to the unique learning trajectories of students~\cite{eaton2010global}.

Recently, LLMs have opened new possibilities for transforming English Education~\cite{caines2023application}. LLMs exhibit remarkable natural language understanding and generation capabilities, making them promising candidates for roles traditionally filled by human tutors. Leveraging LLMs as AI tutors can overcome the inherent limitations of conventional teaching methods, offering scalable, interactive, and personalized learning experiences~\cite{chen2024empowering,schmucker2024ruffle}. Therefore, this position paper argues that \textbf{LLMs can be effective tutors in English education, complementing human expertise and addressing key limitations of traditional methods}.

\begin{figure}[tb!]
\centering
\vspace{0cm}
\includegraphics[scale=0.15]{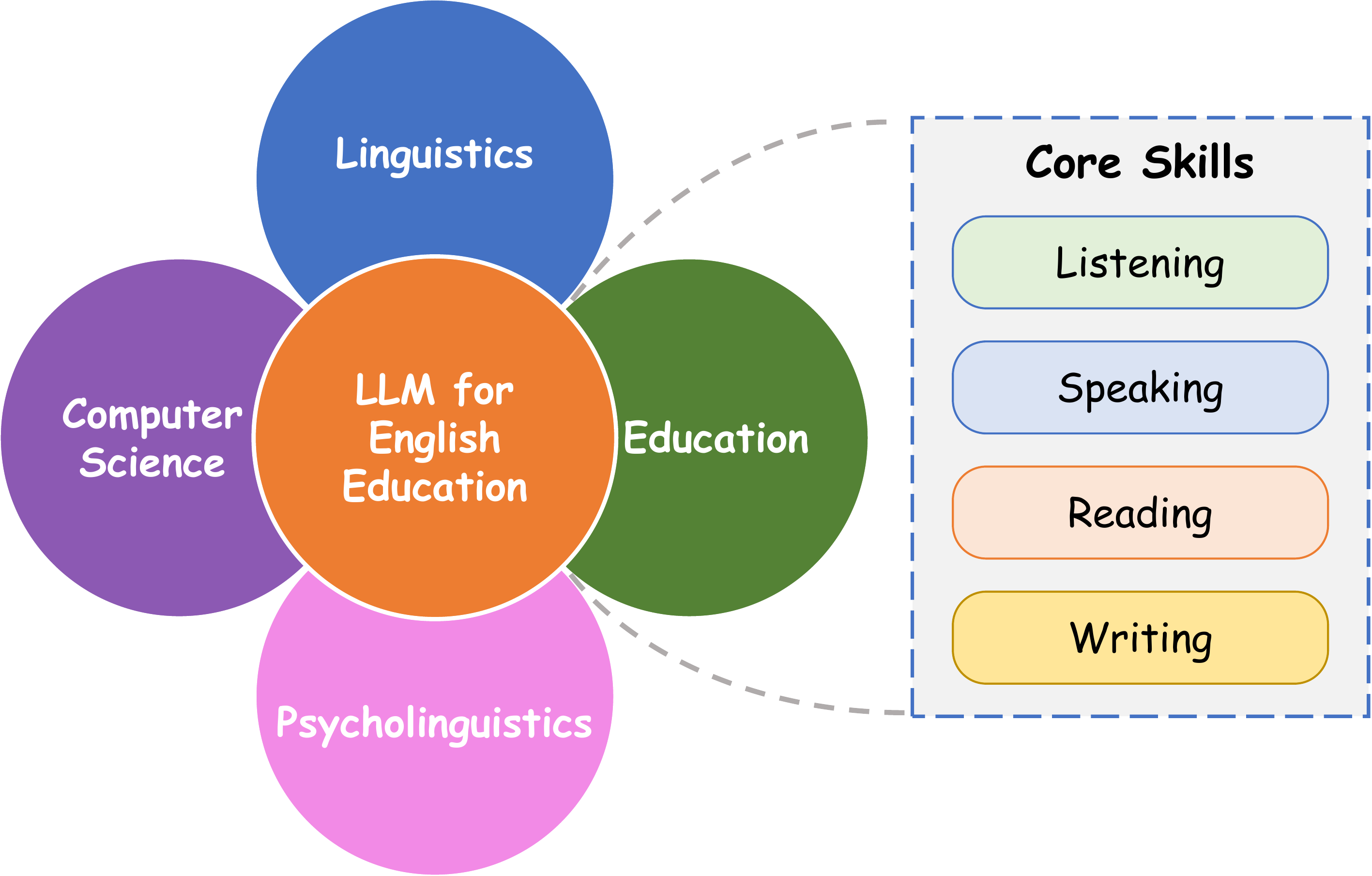}
\caption{Involved disciplines of LLM for English Edu.}
\label{fig:introduction}
\vskip -0.2in
\end{figure}

\begin{figure*}[thb!]
\centering
\vspace{0cm}
\includegraphics[scale=0.26]{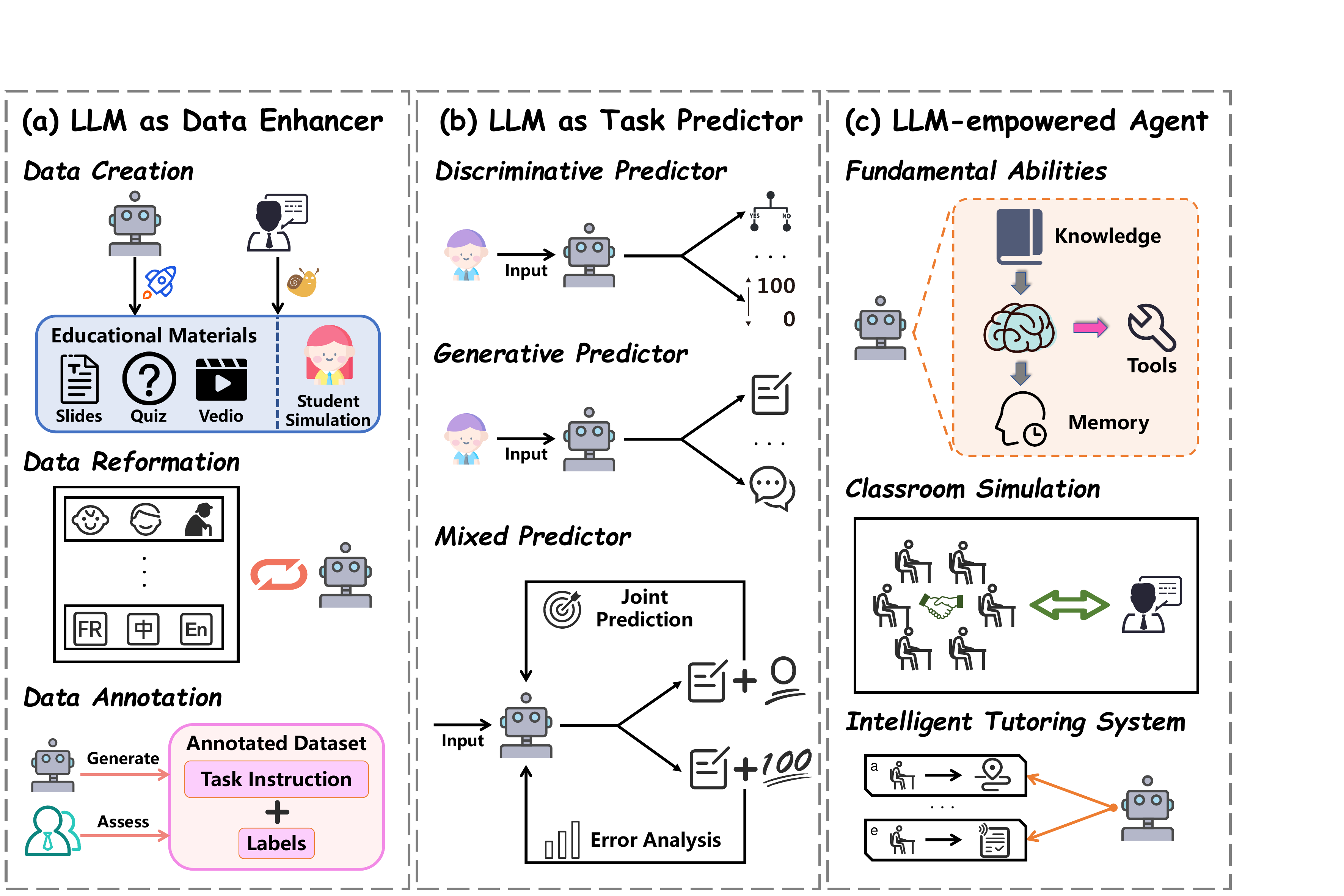}
\caption{Overview of three roles of LLMs in English education. An overview of related literature is provided in Appendix~\ref{app:review}.}
\label{fig:overview}
\end{figure*}

As shown in Figure~\ref{fig:introduction}, English Education intersects with multiple \textit{disciplines}, each of which underscores the potential of LLMs to revolutionize this domain. From the perspective of (1) \textit{computer science}, advancements in machine learning and NLP have enabled LLMs to process and generate human-like language at an unprecedented scale; (2) \textit{linguistics}~\cite{radford2009linguistics} contributes a deeper understanding of grammar, phonetics, and semantics, allowing LLMs to generate accurate and understandable language outputs; (3) \textit{education} provides the foundation for designing effective pedagogical strategies, ensuring that LLMs can deliver personalized, engaging, and developmentally appropriate learning experiences; and finally, (4) \textit{psycholinguistics}~\cite{steinberg2013introduction} bridges the gap between language acquisition and cognitive processes, enabling LLMs to optimize learner interactions by adapting to individual needs and fostering meaningful engagement. Together, these disciplines position LLMs as uniquely capable of addressing the multifaceted challenges of English education.

Moreover, English Education encompasses four \textit{core skills}: listening, speaking, reading, and writing, each of which can be significantly enhanced by LLMs. For listening, LLMs can generate diverse audio materials~\cite{ghosal2023text} and facilitate interactive voice-based exercises, helping learners improve their ability to discern pronunciation, intonation, and contextual meaning. In speaking, LLMs can simulate realistic conversations~\cite{li-etal-2024-eden}, provide pronunciation feedback, and scaffold learners’ oral communication skills through iterative practice. For reading, LLMs can curate leveled texts, generate comprehension questions~\cite{samuel-etal-2024-llms}, and engage learners in discussions that deepen their understanding of written content. Finally, in writing, LLMs can offer real-time grammar, syntax, and style feedback while assisting with idea generation and iterative revisions~\cite{stahl-etal-2024-exploring}. By addressing these core skills holistically, LLMs have the potential to deliver a comprehensive and adaptive learning experience.

Despite these opportunities, the deployment of LLMs in English Education must be approached carefully, ensuring that their integration complements rather than replaces human tutors~\cite{jeon2023large}. As illustrated in Figure~\ref{fig:overview}, this paper explores three critical \textit{roles of LLMs} in this context: their function as \textbf{\textit{data enhancers}} (Section \ref{sec:enhancer}) to optimize learning materials, their capacity as \textbf{\textit{task predictors}} (Section \ref{sec:predictor}) to tailor educational solutions, and their potential as \textbf{\textit{agents}} (Section \ref{sec:agent}) that deliver interactive and adaptive language instruction. By examining these roles, we aim to demonstrate how LLMs can address the limitations of traditional English teaching methods while advancing our understanding of intelligent tutoring systems. Additionally, we discuss potential challenges (Section \ref{sec:challenge}) and future directions (Appendix \ref{app:future_direction}) for integrating LLMs into English Education, offering a technical guideline for researchers and educators to harness their transformative potential. We also describe the paradigm shift of leveraging AI for English Education, starting from the last century, as one of our contributions in Section~\ref{sec:paradigm}.


\section{Background}\label{sec:background}
\subsection{English Education}

Traditional English Education methods often emphasize grammar rules, vocabulary memorization, and repetitive practice, supplemented by limited opportunities for real-world application~\cite{watzke2003lasting}. Such approaches are often constrained by the availability of skilled teachers, the diversity of learners' needs, and the lack of personalized feedback~\cite{williams2004learners}. Recently, many technologies for English Education have been proposed~\cite{alhusaiyan2024systematic}, focusing on solving specific tasks instead of describing the whole picture of English tutoring. While intelligent language tutoring systems have the potential to create adaptive environments, attention to this field is relatively less compared to other subjects like science~\cite{shao2025unlocking} and mathematics~\cite{ahn2024large}. One key reason lies in the inherent complexity of language as an \textit{ill-defined} domain~\cite{schmidt2022artificial}, posing a great challenge in establishing a valid automatic analysis of learner languages due to the vast variability and unpredictability of human language.



\subsection{Large Language Models for Education}
The potential of LLMs in education~\cite{alhafni2024llms}, particularly in English Education~\cite{gao2024exploring,karatacs2024incorporating,cherednichenko2024large}, is immense. Benefiting from large-scale pre-training on extensive corpora, LLMs have demonstrated emergent abilities including (1) \textit{in-context learning}~\cite{dong2022survey}, which allows the model to adapt to new tasks and provide contextually relevant responses based on a few examples provided during the interaction; (2) \textit{instruction following}~\cite{zeng2024evaluating}, which enables the model to process and execute complex user instructions with high accuracy; and (3) \textit{reasoning and planning}~\cite{huang2024understanding}, which allows the model to generate coherent, structured, and context-aware outputs, even for tasks that require multi-step thinking.
However, these fundamental capabilities, while impressive, are insufficient to fully meet the unique demands of English Education. Teaching English requires more than generating grammatically correct sentences or providing accurate translations; it demands a nuanced understanding of pedagogy, learner psychology, and cultural context. \citet{maurya2024unifying} propose an evaluation taxonomy that identifies eight critical dimensions for assessing AI tutors. These dimensions can be broadly categorized into two groups. (1) \textit{Problem-solving abilities} assess the technical capabilities of LLMs to perform tasks relevant to English Education. (2) \textit{Pedagogical alignment abilities} evaluate how well the LLM aligns with effective teaching and learning principles. Pedagogical alignment includes the model's ability to adapt to the learner's proficiency level, provide scaffolded feedback, foster engagement, and maintain motivation. While LLMs can give direct answers, their ability to replicate these nuanced teaching strategies remains a challenge~\cite{wang2024challenges}.



\section{Paradigm Shift}\label{sec:paradigm}

\begin{figure*}[thb!]
\centering
\vspace{0cm}
\includegraphics[scale=0.28]{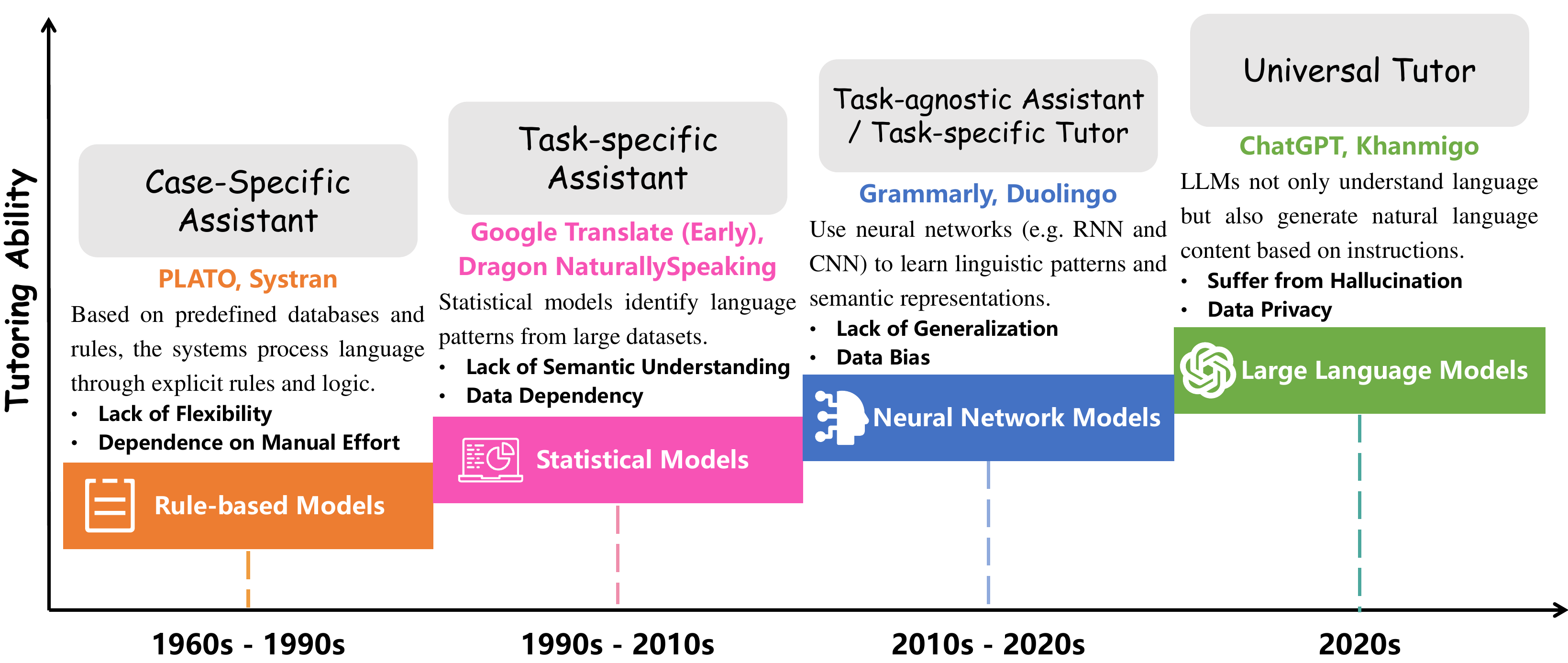}
\caption{Roadmap of English Education.}
\label{fig:paradigm}
\vskip -0.1in
\end{figure*}

The development of AI models for English Education can be broadly categorized into four successive generations as shown in Figure~\ref{fig:paradigm}: (1) \textit{rule-based models}, (2) \textit{statistical models}, (3) \textit{neural models}, and (4) \textit{large language models}.


\paragraph{Stage 1: Rule-based Models (1960s--1990s).}
Early solutions relied on handcrafted linguistic rules to process language in tightly constrained scenarios~\cite{grosan2011rule,c1993towards}. Classical platforms like PLATO~\cite{hart1981language} and Systran~\cite{toma1977systran} operated effectively for highly structured tasks (e.g., grammar drills) but struggled with complex, context-dependent interactions.

\paragraph{Stage 2: Statistical Models (1990s--2010s).}
With the increased availability of digitized corpora, methods such as the early version of Google Translate~\cite{och2006statistical} and Dragon NaturallySpeaking~\cite{blair1997dragon} pioneered statistical pattern mining. These approaches leveraged large datasets to infer linguistic rules and conduct specific tasks probabilistically, improving scalability yet still lacking deeper semantic understanding.

\paragraph{Stage 3: Neural Models (2010s--2020s).}
The advent of deep learning architectures (e.g., RNNs~\cite{yu2019review} and Transformers~\cite{vaswani2017attention}) enabled more robust context modeling, sparking transformative applications like Grammarly~\cite{fitria2021grammarly} and Duolingo~\cite{vesselinov2012duolingo}. These systems offered enhanced personalization and feedback, significantly augmenting learners’ writing and reading comprehension.

\paragraph{Stage 4: Large Language Models (2020s--Present).}
Nowadays, various LLMs (e.g., ChatGPT~\cite{achiam2023gpt}) combine massive pre-training with instruction tuning, achieving impressive results in multi-turn dialogue, individualized scaffolding, and multimodal integration. Tools such as Khanmigo~\cite{anand2023khan} demonstrate LLMs’ potential for real-time conversational practice, dynamic content creation, and inclusive educational support at scale.

\begin{tcolorbox}[top=1pt, bottom=1pt, left=1pt, right=1pt]
\textbf{Our position.}
We foresee next-generation LLMs with deeper alignment to pedagogical principles and stronger guardrails to mitigate misinformation and bias. Future models may integrate multimodal data (e.g., text, image, video, speech) to adapt to diverse learner profiles in real time. These improvements will reinforce the position that LLMs can evolve into more effective tutors for English Education.
\end{tcolorbox}

\section{LLMs as Data Enhancers}\label{sec:enhancer}
Education is a high-stake area where any hallucination could cause devastating harm to humans' cognition activities~\cite{ho2024mitigating}. One of the hallucination causes is from data~\cite{huang2023survey}. Therefore, high-quality and diverse data resources~\cite{long-etal-2024-llms} are critical to ensuring the reliability of incorporating LLMs into English Education. The 1) \textit{creation}, 2) \textit{reformation}, and 3) \textit{annotation} of educational materials are crucial to delivering effective and engaging teaching. Traditional resource development methods often lack the scalability, adaptability, and personalization necessary to meet the diverse needs of learners~\cite{feng-etal-2021-survey,shorten2021text}. In contrast, LLMs emerge as transformative tools capable of enhancing these processes~\cite{wang2024survey,liu2024best}. This section explores how LLMs serve as data enhancers in English Education.

\subsection{Data Creation}
Creating pedagogically sound and learner-specific data is a cornerstone of personalized learning. However, manually creating such resources is time-consuming and often fails to address the wide range of learner needs~\cite{cochran2022improving}. LLMs can revolutionize this process by generating tailored and diverse educational content or responses on demand~\cite{zha2023data,cochran2023improving}.


\paragraph{Educational Materials Generation.} A primary use of LLMs in data creation is the \textit{generation of educational questions} aligned with specific learning objectives. Due to their superior contextual understanding, classic rule-based approaches have largely been eclipsed by neural network-based techniques~\cite{kurdi2020systematic,rathod-etal-2022-educational,mulla2023automatic}. LLMs can produce answer-aware (whose target answer is known) or answer-agnostic (whose answer is open)~\cite{zhang2021review}, resulting in more nuanced exercises and assessments~\cite{xiao-etal-2023-evaluating}.

\paragraph{Student Simulation.}
Simulating the learner’s perspective is crucial for designing adaptive instructional materials. Traditional surveys and standardized tests often fail to capture the complexity of dynamic learner behaviors~\cite{kaser2024simulated}. In contrast, LLM-based approaches enable high-fidelity, context-aware \textit{student simulations}~\cite{liu2024personality,yue2024mathvc}, generating synthetic learners who exhibit realistic mastery levels and evolving behaviors. For instance, \textit{Generative Students}~\cite{lu2024generative} create simulated learners with various competency levels, while \textit{EduAgent}~\cite{xu2024eduagent} integrates cognitive priors to model complex learning trajectories and behaviors better.

\paragraph{Discussion.}
While LLMs excel at generating educational content, current approaches mainly focus on question creation, leaving many areas of English Education underexplored. Essential tasks like generating culturally rich reading materials, context-dependent writing prompts, or dynamic comprehension exercises still lack diversity and depth. Additionally, the student simulations created by LLMs often fail to reflect long-term learning trajectories or the intricacies of individual learning progress.


\subsection{Data Reformation}
In addition to creating new content, LLMs can adapt \textit{existing} materials to better align with current needs. This process, commonly referred to as data reformation, involves (1) changing data types or modalities, (2) paraphrasing materials to match learner proficiency, and (3) enriching raw data with auxiliary signals or contextual content.

\paragraph{Teaching Material Transformation.} Transforming existing materials into different forms can yield more comprehensive and immersive learning experiences. For example, \textit{Book2Dial}\cite{wang-etal-2024-book2dial} generates teacher-student dialogues grounded in textbooks, keeping the content both relevant and informative. Their approach includes multi-turn question generation and answering\cite{kim-etal-2022-generating}, dialogue inpainting~\cite{dai2022dialog}, and role-playing. Likewise, \textit{Slide2Lecture}~\cite{zhang2024awaking} automatically converts lecture slides into structured teaching agendas, enabling interactive follow-up and deeper learner engagement.

\paragraph{Simplification and Paraphrasing.} Another vital application is simplifying or paraphrasing complex texts to specified readability levels~\cite{huang2024generating} without losing key concepts~\cite{al2021automated}. This is particularly beneficial in English Education settings, where language beginners often face advanced vocabulary and complex structures~\cite{day2025evaluating}. Recent advancements in controllable generation~\cite{zhang2023survey} leverage model fine-tuning on curated datasets~\cite{zeng-etal-2023-seen} or decoding-time interventions~\cite{liang2024controllable}, thereby allowing educators to specify text complexity, style, or tone.

\paragraph{Cultural Context Adaptation.}
Beyond linguistic correctness, cultural nuance is another crucial factor in English Education~\cite{byram1989cultural,byram2008foreign}. LLMs can facilitate this process by recontextualizing existing materials to reflect the cultural and social norms of different areas~\cite{liu2024culturally,adilazuarda-etal-2024-towards,kharchenko2024well}. For instance, a short story originally set in an English-speaking environment may be adapted for Japanese students by adjusting the characters’ names, idiomatic expressions, or social customs, while preserving core instructional goals. This cultural adaptation not only enhances learner engagement but also strengthens cross-cultural competencies.

\paragraph{Discussion.}
While LLM-based data reformation can significantly enhance English Education, several gaps warrant attention. Most current studies prioritize textual forms or single-modal approaches, which may overlook valuable \textit{multimodal} resources such as interactive video and audio-based content~\cite{ghosal2023text}. Furthermore, cultural adaptation, although promising, remains underexplored in practical classroom scenarios, particularly for underrepresented persons and culturally sensitive topics. \citet{alkhamissi-etal-2024-investigating} demonstrate how cultural misalignment can increase bias. However, robust empirical \textit{evaluations} are still limited across diverse learners and linguistic backgrounds.


\subsection{Data Annotation}
While \textit{Data Creation} focuses on generating learner-specific data, it often prioritizes diversity and adaptability over precision. The approach is particularly useful for tasks with large label spaces~\cite{ding-etal-2024-data}. In contrast, \textit{Data Annotation} emphasizes producing high-quality, meticulously labeled data that is essential for tasks requiring accuracy and consistency. Unlike data creation, annotated data often undergoes rigorous validation to ensure its accuracy and relevancy~\cite{artemova2024hands}.

\paragraph{Annotation Generation.} LLMs can be central to generating a variety of annotations, including categorical labels, rationales, pedagogical feedback, and linguistic features such as discourse relations. Recent prompt engineering and fine-tuning techniques have further expanded LLMs’ annotation capabilities. For instance, \citet{ye2024excgec} leverage GPT-4 to annotate structured explanations for Chinese grammatical error correction, while \citet{samuel-etal-2024-llms} examine GPT-4 as a surrogate for human annotators in low-resource reading comprehension tasks. Likewise, \citet{li-etal-2024-eden} deploy GPT-4-Turbo for audio transcript annotations. However, inconsistencies across LLMs~\cite{tornberg2024best} remain a serious challenge, posing risks to educational reliability.

\paragraph{Annotation Assessment.}
Although LLM-based annotation is efficient, it also raises critical issues of bias, calibration, and validity, particularly in low-resource language contexts~\cite{bhat-varma-2023-large,jadhav2024limitations}. Automated or semi-automated evaluation strategies have emerged to address these quality concerns. For example, LLMs-as-Judges~\cite{li2024generation,li2024llms,gu2024survey} reduce human overhead by automating evaluation, an approach increasingly explored in education-focused applications~\cite{chiang-etal-2024-large,zhou-etal-2024-llm}. However, purely automated frameworks can still propagate errors or bias.


\paragraph{Discussion.} Although LLMs provide efficient data annotation, the inconsistency across different models remains a critical concern, affecting the quality and reliability of annotated educational materials. These discrepancies hinder the creation of universally reliable educational content, especially in diverse linguistic and cultural contexts. Additionally, automated annotations often lack the nuance needed for pedagogical applications, making it essential to involve human oversight in critical cases to mitigate errors or biases.

\begin{tcolorbox}[top=1pt, bottom=1pt, left=1pt, right=1pt]
\textbf{Our position.}
We acknowledge the current limitations in LLM-based data creation, reformation, and annotation for English Education. However, we believe that with continued interdisciplinary collaboration, these challenges can be addressed. \textit{Future advancements} should focus on enhancing the accuracy and diversity of generated content, improving multi-modal and culturally sensitive learning materials, and integrating more robust systems for human-LLM collaboration~\cite{li-etal-2023-coannotating,wang2024human} in data annotation. This will ensure that LLMs can fully realize their potential as effective tutors in English Education.
\end{tcolorbox}

\section{LLMs as Task Predictors}\label{sec:predictor}
\textit{Task-Based Language Learning (TBLL)}~\cite{nunan1989designing,willis2021framework} as a methodological approach is one of the effective English Education methods. LLMs have demonstrated remarkable capabilities in understanding and generating human language, making them well-suited for addressing numerous tasks in English Education. These tasks can be broadly categorized into three types based on their nature and the role of LLMs: 1) \textit{Discriminative}, 2) \textit{Generative}, and 3) \textit{Mixed} of the above two roles.

\subsection{Discriminative Task Predictors}
Discriminative tasks in English Education primarily involve classifying learner inputs or grading their future performance. Below are some applications that are still calling for improvements:


\paragraph{Automated Assessment.}  The task aims to automatically grade students’ assignments, including essay scoring~\cite{sessler2024can,li2024applying,syamkumar2024improving}, short answer grading~\cite{schneider2023towards,henkel2024can}, and spoken language evaluation~\cite{gao2023investigation,fu2024pronunciation}. LLMs can process learners’ submissions to judge grammar, lexical diversity, coherence, and even spoken fluency, providing instant feedback. This scalability is particularly appealing for large classes, where human evaluators are often overwhelmed and unable to provide timely, personalized critique~\cite{mizumoto2023exploring}.


\paragraph{Knowledge Tracing.}
Given sequences of learning interactions in online learning systems, Knowledge Tracing identifies and tracks students’ evolving mastery of target skills~\cite{shen2024survey,xu2023learning}. LLM-based methods of Knowledge Tracing have been explored in cold-start scenarios~\cite{zhan2024knowledge,jung2024clst}, offering strong generalization by inferring latent learner states from limited data. These approaches can support adaptive learning pathways, giving personalized recommendations based on predicted performance and knowledge gaps.

\paragraph{Discussion.}
Despite their promise in automating and personalizing these discriminative tasks, LLMs still grapple with notable limitations that hinder their utility as robust tutoring tools. First, \emph{misalignment of assessment with expert instructors} poses risks: machine-generated scores may deviate from established rubrics or neglect qualitative nuances, leading to potential discrepancies in grading quality~\cite{kundu2024large}. Second, the \textit{lack of empathy} compounds this issue, as assessments devoid of human judgment risk discouraging learners or overlooking subtle motivational factors~\cite{sharma2024comuniqa}. Knowledge tracing approaches, while promising in cold-start scenarios, struggle with capturing the complexity of long-term learning trajectories and deeper cognitive processes~\cite{cho2024systematic}. These concerns point to the need for more transparent and human-centered methods in utilizing LLMs for assessment.


\subsection{Generative Task Predictors}
Generative tasks involve producing new content or responses. LLMs are known to be adept at these tasks due to their generation capabilities.

\paragraph{Grammatical Error Correction and Explanation.}
In English writing, errors often reveal learners’ gaps in grammar and vocabulary~\cite{hyland2006feedback}. LLMs can detect and correct these errors~\cite{bryant2023grammatical,ye-etal-2023-mixedit}, offering concise explanations~\cite{ye2024excgec} that reinforce language rules. By streamlining error detection and corrections, learners deepen their linguistic understanding.

\paragraph{Feedback Generation.}
Quizzes and exercises remain vital in English Education for practice and targeted remediation~\cite{rashov2024modern}. LLMs enhance this process by delivering prompt, personalized feedback that pinpoints strengths and addresses weaknesses~\cite{borges-etal-2024-teach}. This scalability enables learners to self-regulate and refine their skills without relying solely on human graders~\cite{stamper2024enhancing}.

\paragraph{Socratic Dialogue.}
Moving beyond straightforward Q\&A, Socratic questioning promotes critical thinking and self-reflection~\cite{paul2007critical}. \textit{SocraticLM}~\cite{liusocraticlm}, for example, aligns an LLM with open-ended, inquiry-based teaching principles, guiding learners through iterative exploration rather than prescriptive correction. In theory, this fosters deeper conceptual understanding and active learner engagement.

\paragraph{Discussion.}
Despite the promise of LLM-based generation in English Education, multiple uncertainties persist. \textit{Determining how to provide automatic feedback that genuinely maximizes learning outcomes} is an ongoing challenge~\cite{stamper2024enhancing}, particularly given education’s risk-averse culture and high accountability standards~\cite{xiao2024humanaicollaborativeessayscoring}. Moreover, while LLMs like SocraticLM have demonstrated success in domains like mathematics, their applicability to English Education contexts has not been thoroughly validated~\cite{liusocraticlm}. As such, the design of strategies and follow-up queries remains an open question in ensuring these systems track and respond to learners' cognitive states.

\subsection{Mixed Task Predictors}
Mixed tasks integrate discriminative and generative elements, requiring LLMs to evaluate learner inputs and generate meaningful feedback or suggestions. These tasks are particularly valuable in fostering an interactive and adaptive learning experience, as they bridge the gap between evaluation and instruction.

\paragraph{Automated Assessment with Feedback.}
While discriminative systems for automated essay scoring and speech evaluation primarily focus on assigning grades, LLMs extend these capabilities by simultaneously generating formative feedback~\cite{katuka2024investigating,stahl2024exploring}. For example, an LLM can evaluate the coherence and lexical diversity of a written essay, then offer specific revision strategies. In speaking practice, it can measure fluency and pronunciation accuracy while suggesting drills to refine intonation or stress patterns. Through this combination of scoring and tailored advice, learners gain a deeper understanding of their strengths and areas for improvement.

\paragraph{Error Analysis.} Error Analysis systematically uncovers and categorizes learners’ missteps, from syntactic lapses in writing to flawed pronunciations in speaking~\cite{james2013errors,erdougan2005contribution}. LLMs functioning in a mixed capacity can classify these errors and generate corrective guidance, providing revised sentences, clarifications of grammatical rules, or remediation exercises for identified weaknesses~\cite{myles2002second,mashoor2020error}. Such insight facilitates targeted interventions that enhance language proficiency across modalities, including reading and listening.


\paragraph{Discussion.} Mixed-task systems hold promise by combining assessment and feedback generation, but they face notable challenges. One major issue is the \textit{weak alignment} between scoring mechanisms and the quality of feedback provided~\cite{stahl2024exploring}. For example, while essay scoring systems may deliver comprehensive evaluations, the feedback often lacks specificity, limiting its instructional value. Additionally, although error analysis has potential, \textit{the absence of standardized pedagogical benchmarks}, especially in oral tasks, hampers the reliability and comparability of LLM-based tools~\cite{leu1982oral}.

\begin{tcolorbox}[top=1pt, bottom=1pt, left=1pt, right=1pt]
\textbf{Our position.} While LLMs offer scalable solutions for task prediction in English Education, their current limitations—such as misalignment with expert assessments, lack of empathy, and weak alignment between assessment and feedback—require ongoing refinement. \textit{Future research} should focus on improving model transparency, enhancing the cultural and emotional sensitivity of LLMs, and refining task predictors to better reflect long-term learning trajectories and learner motivation. Additionally, developing standardized pedagogical benchmarks for error analysis will help ensure the consistency and reliability of LLM-generated feedback.
\end{tcolorbox}

\section{LLM-empowered Agent}\label{sec:agent}
In this section, we delve into the potential of LLMs as intelligent tutoring agents in English Education. LLMs can act as catalysts for personalized learning, addressing the long-standing scalability, adaptability, and inclusivity challenges in traditional teaching paradigms.

\subsection{Fundamental Abilities}
This section highlights five key abilities of LLM-empowered agents that enable them to function as adaptive tutors.

\paragraph{Knowledge Integration.} LLMs excel at merging structured educational knowledge graphs~\cite{abu2024knowledge,hu2024foke} with unstructured textual data~\cite{li2024supporting,modran2024llm}, providing rich, contextualized information on linguistic constructs and cultural nuances. Their ability to perform real-time knowledge editing~\cite{wang2024knowledge,zhang2024comprehensive} ensures learners receive content aligned with evolving language usage, addressing the inherent limitations of static materials.

\paragraph{Pedagogical Alignment.} LLMs require embedding with pedagogical principles to facilitate genuine learning experiences~\cite{carroll1965contributions,taneja1995educational}. Recent work incorporates theoretical frameworks, such as Bloom’s taxonomy~\cite{bloom1956taxonomy}, to guide LLMs in systematically addressing different cognitive levels~\cite{jiang2024llms}. Approaches like \textit{Pedagogical Chain of Thought}~\cite{jiang2024llms} and \textit{preference learning}~\cite{sonkar-etal-2024-pedagogical,rafailov2024direct} focus on aligning model responses with educational objectives.


\paragraph{Planning.} By assisting in crafting teaching objectives and lesson designs, LLMs can handle complex tasks such as differentiated instruction~\cite{hu2024teaching}. LessonPlanner~\cite{fan2024lessonplanner} has been proposed to assist novice teachers in preparing lesson plans, with expert interviews confirming its effectiveness. \citet{zheng2024automatic} propose a three-stage process to produce customized lesson plans, using Retrieval-Augmented Generation (RAG), self-critique, and subsequent refinement.

\paragraph{Memory.} Effective tutoring systems track learner histories and tailor subsequent interactions accordingly~\cite{jiang2024ai,chen2024empowering}. When serving as memory-augmented agents, LLMs can retain individualized data—such as repeated grammar mistakes or overlooked vocabulary—thereby improving continuity and enabling consistent scaffolding of future learning tasks.

\paragraph{Tool Using.} Beyond textual interactions, LLM-based agents can integrate specialized tools to streamline the educational ecosystem, from cognitive diagnosis modules~\cite{ma2019cognitive} to report generators~\cite{zhou2025study}. By orchestrating these resources, LLMs seamlessly unify diverse utilities under a single interface, enhancing learner experience and instructional efficiency.


\subsection{Applications}
Although still in its early stages, LLM-empowered agents have already started to show promising applications in English Education.

\paragraph{Classroom Simulation.}
Classroom simulation leverages LLM-empowered agents to recreate complex, interactive learning settings without the logistical hurdles of organizing physical classrooms~\cite{zhang2024simulating}. By simulating virtual students and tutors, researchers can study pedagogical strategies at scale, generate diverse learner interactions, and refine teaching techniques. Moreover, this virtual data can be used to fine-tune LLMs for specific educational contexts and learner profiles~\cite{liusocraticlm}, offering a cost-effective and adaptable approach to language instruction.

\paragraph{Intelligent Tutoring System (ITS).}
LLM-based agents have demonstrated the capacity to provide dynamic, personalized tutoring experiences~\cite{wang2025llm,kwon-etal-2024-biped}, effectively identifying learner weaknesses through large-scale linguistic analysis~\cite{caines2023application}. This makes them promising for delivering individualized instruction at scale. Although current ITS applications in mathematics~\cite{pal2024autotutor} and science~\cite{stamper2024enhancing} have shown success, the extension to English Education requires nuanced handling of cultural and contextual elements, as well as the unpredictability of human language usage.

\paragraph{Discussion.}
Despite the promise of these applications, critical challenges remain. Existing classroom simulation frameworks often \textit{lack standardized benchmarks for English Education}, making it difficult to assess the efficacy and generalizability of developed systems~\cite{zhang2024simulating}. In addition, evaluating language-specific tutoring strategies, including real-time conversational practice and holistic skill integration, remains an underexplored frontier. Addressing these gaps requires \textit{new datasets and metrics} centered on holistic skill development and interdisciplinary collaboration.

\begin{tcolorbox}[top=1pt, bottom=1pt, left=1pt, right=1pt]
\textbf{Our position.}
We argue that \textit{future research} should focus on integrating multimodal learning tasks~\cite{sonlu2024effects} and developing standardized frameworks for evaluating English Education simulations. Moreover, LLMs should evolve beyond text-based capabilities to provide real-time, context-sensitive feedback, particularly in speaking and listening. Interdisciplinary collaboration and the creation of new datasets tailored to English Education are crucial for refining these systems and ensuring their scalability and inclusivity in language instruction.

\end{tcolorbox}



\section{Challenges}
\label{sec:challenge}
While we posit that LLMs have the potential to revolutionize English Education, realizing their full promise requires addressing key challenges. This section offers a concise overview of these challenges, followed by directions that could guide future research and deployment.

\paragraph{Ensuring Reliability and Mitigating Hallucinations.}
LLMs may produce hallucinations~\cite{huang2023survey} that can mislead learners and undermine pedagogical goals. This risk intensifies in high-stakes educational environments, where trust and correctness are paramount. Future directions include enhancing data quality and diversity for training~\cite{long-etal-2024-llms}, developing techniques to integrate LLM outputs with structured domain knowledge and pedagogical rules, and employing rigorous automated and human-in-the-loop validation mechanisms to minimize such detrimental outcomes and improve the factual grounding of LLM-generated educational content.

\paragraph{Addressing Bias and Ethical Considerations.}
As LLMs inherit biases from their training data, these systems may produce culturally insensitive, stereotypical, or unfair responses, potentially harming students from diverse linguistic and sociocultural backgrounds. Moreover, significant privacy concerns emerge when collecting and using learner data to personalize instruction, particularly for K-12 students. Future research must focus on developing robust governance frameworks, transparent documentation of data sources and model behaviors, and advanced bias detection and mitigation strategies~\cite{borah2024towards,he2024mitigating} to ensure that LLM-based tools for English Education are equitable, fair, and uphold stringent data protection standards.

\paragraph{Aligning With Pedagogical Principles.}
LLMs excel at generating fluent language but often lack deep pedagogical alignment, particularly for tasks requiring developmental sensitivity, learner motivation strategies, or differentiated instruction tailored to individual learning needs. Their general-purpose nature means they do not inherently account for established language acquisition theories or specific curricular standards~\cite{razafinirina2024pedagogical}. A crucial future direction is the development of methodologies to better imbue LLMs with pedagogical intelligence. This includes co-designing LLM applications with educators, fine-tuning models on high-quality pedagogical interaction data, and creating architectures that can dynamically adapt to learners’ cognitive states and developmental needs in English language learning.

\section{Conclusion}
This paper emphasizes the transformative potential of LLMs in English Education, positioning them as valuable tutors to complement traditional teaching methods. Through their roles as data enhancers, task predictors, and agents, LLMs can provide adaptive learning experiences across the core skills of listening, speaking, reading, and writing. This paper encourages continuing dialogue and interdisciplinary collaboration to responsibly integrate LLMs into educational ecosystems.

\section*{Limitations}
\paragraph{Emphasis on potential over practical implementation barriers.}
This paper primarily focuses on the potential of LLMs to serve as effective tutors in English Education, outlining beneficial roles as data enhancers, task predictors, and agents. While we acknowledge the existence of challenges (to be discussed in Appendix~\ref{sec:challenge}), a limitation of this position is that the main arguments may not fully capture the considerable practical, socio-economic, and infrastructural hurdles that could impede the equitable and effective implementation of these LLM roles across diverse global educational contexts and resource settings.

\paragraph{Generalizability and contextual adaptation of proposed roles.}
We propose three broad roles for LLMs in English Education. However, this paper does not provide an exhaustive analysis of how the efficacy and suitability of LLMs in these roles might vary significantly across different target languages (especially low-resource languages), specific learner demographics (e.g., preschoolers vs. K-12 vs. adult learners, learners with disabilities), diverse cultural contexts, or varying pedagogical philosophies. The general framework presented may require substantial adaptation and further research to be effectively applied in specific English Education scenarios.

\paragraph{Nuances of human-LLM pedagogical interaction.}
While advocating for LLMs as tutors that can complement human expertise, this position paper does not delve deeply into the complex dynamics of the pedagogical interactions between learners, LLM-based tutors, human educators, and parents. Critical aspects such as optimizing the collaborative model, designing effective training for educators to leverage LLMs, mitigating risks of learner over-reliance, and ensuring that LLM interactions foster deep learning rather than superficial engagement are multifaceted issues that warrant more extensive investigation than afforded by the scope of this paper.



\section*{Acknowledgements}
This research is supported in part by NSF under grants III-2106758, and POSE-2346158. This research is also supported by National Natural Science Foundation of China (Grant No.62276154); Research Center for Computer Network (Shenzhen) Ministry of Education, the Natural Science Foundation of Guangdong Province (Grant No.2023A1515012914 and 440300241033100801770); Basic Research Fund of Shenzhen City (Grant No.JCYJ20210324120012033, JCYJ20240813112009013 and GJHZ20240218113603006); The Major Key Project of PCL for Experiments and Applications (PCL2024A08).

\bibliography{main}

\begin{thebibliography}{187}
\providecommand{\natexlab}[1]{#1}

\bibitem[{Abu-Rasheed et~al.(2024)Abu-Rasheed, Weber, and Fathi}]{abu2024knowledge}
Hasan Abu-Rasheed, Christian Weber, and Madjid Fathi. 2024.
\newblock Knowledge graphs as context sources for llm-based explanations of learning recommendations.
\newblock \emph{arXiv preprint arXiv:2403.03008}.

\bibitem[{Achiam et~al.(2023)Achiam, Adler, Agarwal, Ahmad, Akkaya, Aleman, Almeida, Altenschmidt, Altman, Anadkat et~al.}]{achiam2023gpt}
Josh Achiam, Steven Adler, Sandhini Agarwal, Lama Ahmad, Ilge Akkaya, Florencia~Leoni Aleman, Diogo Almeida, Janko Altenschmidt, Sam Altman, Shyamal Anadkat, and 1 others. 2023.
\newblock Gpt-4 technical report.
\newblock \emph{arXiv preprint arXiv:2303.08774}.

\bibitem[{Adilazuarda et~al.(2024)Adilazuarda, Mukherjee, Lavania, Singh, Aji, O{'}Neill, Modi, and Choudhury}]{adilazuarda-etal-2024-towards}
Muhammad~Farid Adilazuarda, Sagnik Mukherjee, Pradhyumna Lavania, Siddhant~Shivdutt Singh, Alham~Fikri Aji, Jacki O{'}Neill, Ashutosh Modi, and Monojit Choudhury. 2024.
\newblock \href {https://doi.org/10.18653/v1/2024.emnlp-main.882} {Towards measuring and modeling {\textquotedblleft}culture{\textquotedblright} in {LLM}s: A survey}.
\newblock In \emph{Proceedings of the 2024 Conference on Empirical Methods in Natural Language Processing}, pages 15763--15784, Miami, Florida, USA. Association for Computational Linguistics.

\bibitem[{Ahn et~al.(2024)Ahn, Verma, Lou, Liu, Zhang, and Yin}]{ahn2024large}
Janice Ahn, Rishu Verma, Renze Lou, Di~Liu, Rui Zhang, and Wenpeng Yin. 2024.
\newblock Large language models for mathematical reasoning: Progresses and challenges.
\newblock \emph{arXiv preprint arXiv:2402.00157}.

\bibitem[{Al-Thanyyan and Azmi(2021)}]{al2021automated}
Suha~S Al-Thanyyan and Aqil~M Azmi. 2021.
\newblock Automated text simplification: a survey.
\newblock \emph{ACM Computing Surveys (CSUR)}, 54(2):1--36.

\bibitem[{Alhafni et~al.(2024)Alhafni, Vajjala, Bann{\`o}, Maurya, and Kochmar}]{alhafni2024llms}
Bashar Alhafni, Sowmya Vajjala, Stefano Bann{\`o}, Kaushal~Kumar Maurya, and Ekaterina Kochmar. 2024.
\newblock Llms in education: Novel perspectives, challenges, and opportunities.
\newblock \emph{arXiv preprint arXiv:2409.11917}.

\bibitem[{Alhusaiyan(2024)}]{alhusaiyan2024systematic}
Eman Alhusaiyan. 2024.
\newblock A systematic review of current trends in artificial intelligence in foreign language learning.
\newblock \emph{Saudi Journal of Language Studies}.

\bibitem[{Alhusaiyan(2025)}]{alhusaiyan2025systematic}
Eman Alhusaiyan. 2025.
\newblock A systematic review of current trends in artificial intelligence in foreign language learning.
\newblock \emph{Saudi Journal of Language Studies}, 5(1):1--16.

\bibitem[{AlKhamissi et~al.(2024)AlKhamissi, ElNokrashy, Alkhamissi, and Diab}]{alkhamissi-etal-2024-investigating}
Badr AlKhamissi, Muhammad ElNokrashy, Mai Alkhamissi, and Mona Diab. 2024.
\newblock \href {https://doi.org/10.18653/v1/2024.acl-long.671} {Investigating cultural alignment of large language models}.
\newblock In \emph{Proceedings of the 62nd Annual Meeting of the Association for Computational Linguistics (Volume 1: Long Papers)}, pages 12404--12422, Bangkok, Thailand. Association for Computational Linguistics.

\bibitem[{Anand(2023)}]{anand2023khan}
Preeti Anand. 2023.
\newblock Khan academy creates gpt-4 based helper khanmigo marking formal entry of ai into education.

\bibitem[{Artemova et~al.(2024)Artemova, Tsvigun, Schlechtweg, Fedorova, Tilga, and Obmoroshev}]{artemova2024hands}
Ekaterina Artemova, Akim Tsvigun, Dominik Schlechtweg, Natalia Fedorova, Sergei Tilga, and Boris Obmoroshev. 2024.
\newblock Hands-on tutorial: Labeling with llm and human-in-the-loop.
\newblock \emph{arXiv preprint arXiv:2411.04637}.

\bibitem[{Asthana et~al.(2024)Asthana, Rashkin, Clark, Huot, and Lapata}]{asthana-etal-2024-evaluating}
Sumit Asthana, Hannah Rashkin, Elizabeth Clark, Fantine Huot, and Mirella Lapata. 2024.
\newblock \href {https://doi.org/10.18653/v1/2024.emnlp-main.357} {Evaluating {LLM}s for targeted concept simplification for domain-specific texts}.
\newblock In \emph{Proceedings of the 2024 Conference on Empirical Methods in Natural Language Processing}, pages 6208--6226, Miami, Florida, USA. Association for Computational Linguistics.

\bibitem[{Bhat and Varma(2023)}]{bhat-varma-2023-large}
Savita Bhat and Vasudeva Varma. 2023.
\newblock \href {https://doi.org/10.18653/v1/2023.eval4nlp-1.8} {Large language models as annotators: A preliminary evaluation for annotating low-resource language content}.
\newblock In \emph{Proceedings of the 4th Workshop on Evaluation and Comparison of NLP Systems}, pages 100--107, Bali, Indonesia. Association for Computational Linguistics.

\bibitem[{Blair(1997)}]{blair1997dragon}
Christopher Blair. 1997.
\newblock Dragon--naturallyspeaking.
\newblock \emph{Journal of Osteopathic Medicine}, 97(12):711--711.

\bibitem[{Bloom et~al.(1956)Bloom, Engelhart, Furst, Hill, Krathwohl et~al.}]{bloom1956taxonomy}
Benjamin~S Bloom, Max~D Engelhart, Edward~J Furst, Walker~H Hill, David~R Krathwohl, and 1 others. 1956.
\newblock \emph{Taxonomy of educational objectives: The classification of educational goals. Handbook 1: Cognitive domain}.
\newblock Longman New York.

\bibitem[{Borah and Mihalcea(2024)}]{borah2024towards}
Angana Borah and Rada Mihalcea. 2024.
\newblock Towards implicit bias detection and mitigation in multi-agent llm interactions.
\newblock \emph{arXiv preprint arXiv:2410.02584}.

\bibitem[{Borges et~al.(2024)Borges, Tandon, K{\"a}ser, and Bosselut}]{borges-etal-2024-teach}
Beatriz Borges, Niket Tandon, Tanja K{\"a}ser, and Antoine Bosselut. 2024.
\newblock \href {https://doi.org/10.18653/v1/2024.emnlp-main.674} {Let me teach you: Pedagogical foundations of feedback for language models}.
\newblock In \emph{Proceedings of the 2024 Conference on Empirical Methods in Natural Language Processing}, pages 12082--12104, Miami, Florida, USA. Association for Computational Linguistics.

\bibitem[{Bryant et~al.(2023)Bryant, Yuan, Qorib, Cao, Ng, and Briscoe}]{bryant2023grammatical}
Christopher Bryant, Zheng Yuan, Muhammad~Reza Qorib, Hannan Cao, Hwee~Tou Ng, and Ted Briscoe. 2023.
\newblock Grammatical error correction: A survey of the state of the art.
\newblock \emph{Computational Linguistics}, 49(3):643--701.

\bibitem[{Byram(1989)}]{byram1989cultural}
M~Byram. 1989.
\newblock Cultural studies in foreign language education.
\newblock \emph{Multilingual Matters}, 61.

\bibitem[{Byram(2008)}]{byram2008foreign}
Michael Byram. 2008.
\newblock \emph{From foreign language education to education for intercultural citizenship: Essays and reflections}, volume~17.
\newblock Multilingual matters.

\bibitem[{C~Angelides and Garcia(1993)}]{c1993towards}
Marios C~Angelides and Isabel Garcia. 1993.
\newblock Towards an intelligent knowledge based tutoring system for foreign language learning.
\newblock \emph{Journal of computing and information technology}, 1(1):15--28.

\bibitem[{Caines et~al.(2023)Caines, Benedetto, Taslimipoor, Davis, Gao, Andersen, Yuan, Elliott, Moore, Bryant et~al.}]{caines2023application}
Andrew Caines, Luca Benedetto, Shiva Taslimipoor, Christopher Davis, Yuan Gao, Oeistein Andersen, Zheng Yuan, Mark Elliott, Russell Moore, Christopher Bryant, and 1 others. 2023.
\newblock On the application of large language models for language teaching and assessment technology.
\newblock \emph{arXiv preprint arXiv:2307.08393}.

\bibitem[{Carroll(1965)}]{carroll1965contributions}
John~B Carroll. 1965.
\newblock The contributions of psychological theory and educational research to the teaching of foreign languages.
\newblock \emph{The modern language journal}, 49(5):273--281.

\bibitem[{Chen et~al.(2024)Chen, Ding, Zheng, Liu, Sun, and Zhou}]{chen2024empowering}
Yulin Chen, Ning Ding, Hai-Tao Zheng, Zhiyuan Liu, Maosong Sun, and Bowen Zhou. 2024.
\newblock Empowering private tutoring by chaining large language models.
\newblock In \emph{Proceedings of the 33rd ACM International Conference on Information and Knowledge Management}, pages 354--364.

\bibitem[{Cherednichenko et~al.(2024)Cherednichenko, Yanholenko, Badan, Onishchenko, and Akopiants}]{cherednichenko2024large}
Olga Cherednichenko, Olha Yanholenko, Antonina Badan, Nataliia Onishchenko, and Nunu Akopiants. 2024.
\newblock Large language models for foreign language acquisition.

\bibitem[{Chiang et~al.(2024)Chiang, Chen, Kuan, Yang, and Lee}]{chiang-etal-2024-large}
Cheng-Han Chiang, Wei-Chih Chen, Chun-Yi Kuan, Chienchou Yang, and Hung-yi Lee. 2024.
\newblock \href {https://doi.org/10.18653/v1/2024.emnlp-main.146} {Large language model as an assignment evaluator: Insights, feedback, and challenges in a 1000+ student course}.
\newblock In \emph{Proceedings of the 2024 Conference on Empirical Methods in Natural Language Processing}, pages 2489--2513, Miami, Florida, USA. Association for Computational Linguistics.

\bibitem[{Cho et~al.(2024)Cho, AlMamlook, and Gharaibeh}]{cho2024systematic}
Yongwan Cho, Rabia~Emhamed AlMamlook, and Tasnim Gharaibeh. 2024.
\newblock A systematic review of knowledge tracing and large language models in education: Opportunities, issues, and future research.
\newblock \emph{arXiv preprint arXiv:2412.09248}.

\bibitem[{Cochran et~al.(2022)Cochran, Cohn, Hutchins, Biswas, and Hastings}]{cochran2022improving}
Keith Cochran, Clayton Cohn, Nicole Hutchins, Gautam Biswas, and Peter Hastings. 2022.
\newblock Improving automated evaluation of formative assessments with text data augmentation.
\newblock In \emph{International Conference on Artificial Intelligence in Education}, pages 390--401. Springer.

\bibitem[{Cochran et~al.(2023)Cochran, Cohn, Rouet, and Hastings}]{cochran2023improving}
Keith Cochran, Clayton Cohn, Jean~Francois Rouet, and Peter Hastings. 2023.
\newblock Improving automated evaluation of student text responses using gpt-3.5 for text data augmentation.
\newblock In \emph{International Conference on Artificial Intelligence in Education}, pages 217--228. Springer.

\bibitem[{Dai et~al.(2022)Dai, Chaganty, Zhao, Amini, Rashid, Green, and Guu}]{dai2022dialog}
Zhuyun Dai, Arun~Tejasvi Chaganty, Vincent~Y Zhao, Aida Amini, Qazi~Mamunur Rashid, Mike Green, and Kelvin Guu. 2022.
\newblock Dialog inpainting: Turning documents into dialogs.
\newblock In \emph{International conference on machine learning}, pages 4558--4586. PMLR.

\bibitem[{Day et~al.(2025)Day, Cirica, Clapp, Penkova, Giroux, Banta, Bordeau, Mutteneni, and Sawyer}]{day2025evaluating}
Stephanie~L. Day, Jacapo Cirica, Steven~R. Clapp, Veronika Penkova, Amy~E. Giroux, Abbey Banta, Catherine Bordeau, Poojitha Mutteneni, and Ben~D. Sawyer. 2025.
\newblock \href {https://arxiv.org/abs/2501.09158} {Evaluating genai for simplifying texts for education: Improving accuracy and consistency for enhanced readability}.
\newblock \emph{Preprint}, arXiv:2501.09158.

\bibitem[{Ding et~al.(2024)Ding, Qin, Zhao, Luo, Li, Chen, Xia, Hu, Luu, and Joty}]{ding-etal-2024-data}
Bosheng Ding, Chengwei Qin, Ruochen Zhao, Tianze Luo, Xinze Li, Guizhen Chen, Wenhan Xia, Junjie Hu, Anh~Tuan Luu, and Shafiq Joty. 2024.
\newblock \href {https://doi.org/10.18653/v1/2024.findings-acl.97} {Data augmentation using {LLM}s: Data perspectives, learning paradigms and challenges}.
\newblock In \emph{Findings of the Association for Computational Linguistics: ACL 2024}, pages 1679--1705, Bangkok, Thailand. Association for Computational Linguistics.

\bibitem[{Dong et~al.(2022)Dong, Li, Dai, Zheng, Ma, Li, Xia, Xu, Wu, Liu et~al.}]{dong2022survey}
Qingxiu Dong, Lei Li, Damai Dai, Ce~Zheng, Jingyuan Ma, Rui Li, Heming Xia, Jingjing Xu, Zhiyong Wu, Tianyu Liu, and 1 others. 2022.
\newblock A survey on in-context learning.
\newblock \emph{arXiv preprint arXiv:2301.00234}.

\bibitem[{Eaton(2010)}]{eaton2010global}
Sarah~Elaine Eaton. 2010.
\newblock \emph{Global Trends in Language Learning in the 21st Century.}
\newblock ERIC.

\bibitem[{Ehrenberg et~al.(2001)Ehrenberg, Brewer, Gamoran, and Willms}]{ehrenberg2001class}
Ronald~G Ehrenberg, Dominic~J Brewer, Adam Gamoran, and J~Douglas Willms. 2001.
\newblock Class size and student achievement.
\newblock \emph{Psychological science in the public interest}, 2(1):1--30.

\bibitem[{Erdo{\u{g}}an(2005)}]{erdougan2005contribution}
Vacide Erdo{\u{g}}an. 2005.
\newblock Contribution of error analysis to foreign language teaching.
\newblock \emph{Mersin {\"U}niversitesi E{\u{g}}itim Fak{\"u}ltesi Dergisi}, 1(2).

\bibitem[{Fan et~al.(2024)Fan, Chen, Wang, and Peng}]{fan2024lessonplanner}
Haoxiang Fan, Guanzheng Chen, Xingbo Wang, and Zhenhui Peng. 2024.
\newblock Lessonplanner: Assisting novice teachers to prepare pedagogy-driven lesson plans with large language models.
\newblock In \emph{Proceedings of the 37th Annual ACM Symposium on User Interface Software and Technology}, pages 1--20.

\bibitem[{Fang et~al.(2023)Fang, Yang, Lan, Wong, Hu, Chao, and Zhang}]{fang2023chatgpt}
Tao Fang, Shu Yang, Kaixin Lan, Derek~F Wong, Jinpeng Hu, Lidia~S Chao, and Yue Zhang. 2023.
\newblock Is chatgpt a highly fluent grammatical error correction system? a comprehensive evaluation.
\newblock \emph{arXiv preprint arXiv:2304.01746}.

\bibitem[{Favero et~al.(2024)Favero, P{\'e}rez-Ortiz, K{\"a}ser, and Oliver}]{favero2024enhancing}
Lucile Favero, Juan~Antonio P{\'e}rez-Ortiz, Tanja K{\"a}ser, and Nuria Oliver. 2024.
\newblock Enhancing critical thinking in education by means of a socratic chatbot.
\newblock \emph{arXiv preprint arXiv:2409.05511}.

\bibitem[{Fei et~al.(2023)Fei, Cui, Yang, Lam, Lan, and Shi}]{fei-etal-2023-enhancing}
Yuejiao Fei, Leyang Cui, Sen Yang, Wai Lam, Zhenzhong Lan, and Shuming Shi. 2023.
\newblock \href {https://doi.org/10.18653/v1/2023.acl-long.413} {Enhancing grammatical error correction systems with explanations}.
\newblock In \emph{Proceedings of the 61st Annual Meeting of the Association for Computational Linguistics (Volume 1: Long Papers)}, pages 7489--7501, Toronto, Canada. Association for Computational Linguistics.

\bibitem[{Feng et~al.(2021)Feng, Gangal, Wei, Chandar, Vosoughi, Mitamura, and Hovy}]{feng-etal-2021-survey}
Steven~Y. Feng, Varun Gangal, Jason Wei, Sarath Chandar, Soroush Vosoughi, Teruko Mitamura, and Eduard Hovy. 2021.
\newblock \href {https://doi.org/10.18653/v1/2021.findings-acl.84} {A survey of data augmentation approaches for {NLP}}.
\newblock In \emph{Findings of the Association for Computational Linguistics: ACL-IJCNLP 2021}, pages 968--988, Online. Association for Computational Linguistics.

\bibitem[{Fitria(2021)}]{fitria2021grammarly}
Tira~Nur Fitria. 2021.
\newblock Grammarly as ai-powered english writing assistant: Students’ alternative for writing english.
\newblock \emph{Metathesis: Journal of English Language, Literature, and Teaching}, 5(1):65--78.

\bibitem[{Freyer et~al.(2024)Freyer, Kempt, and Kl{\"o}ser}]{freyer2024easy}
Nils Freyer, Hendrik Kempt, and Lars Kl{\"o}ser. 2024.
\newblock Easy-read and large language models: on the ethical dimensions of llm-based text simplification.
\newblock \emph{Ethics and Information Technology}, 26(3):50.

\bibitem[{Fu et~al.(2024)Fu, Peng, Yang, and Zhou}]{fu2024pronunciation}
Kaiqi Fu, Linkai Peng, Nan Yang, and Shuran Zhou. 2024.
\newblock Pronunciation assessment with multi-modal large language models.
\newblock \emph{arXiv preprint arXiv:2407.09209}.

\bibitem[{Gao et~al.(2024)Gao, Wang, and Wang}]{gao2024exploring}
Yang Gao, Qikai Wang, and Xiaochen Wang. 2024.
\newblock Exploring efl university teachers’ beliefs in integrating chatgpt and other large language models in language education: a study in china.
\newblock \emph{Asia Pacific Journal of Education}, 44(1):29--44.

\bibitem[{Gao et~al.(2023)Gao, Nuchged, Li, and Peng}]{gao2023investigation}
Yingming Gao, Baorian Nuchged, Ya~Li, and Linkai Peng. 2023.
\newblock An investigation of applying large language models to spoken language learning.
\newblock \emph{Applied Sciences}, 14(1):224.

\bibitem[{Ghosal et~al.(2023)Ghosal, Majumder, Mehrish, and Poria}]{ghosal2023text}
Deepanway Ghosal, Navonil Majumder, Ambuj Mehrish, and Soujanya Poria. 2023.
\newblock Text-to-audio generation using instruction-tuned llm and latent diffusion model.
\newblock \emph{arXiv preprint arXiv:2304.13731}.

\bibitem[{Grosan et~al.(2011)Grosan, Abraham, Grosan, and Abraham}]{grosan2011rule}
Crina Grosan, Ajith Abraham, Crina Grosan, and Ajith Abraham. 2011.
\newblock Rule-based expert systems.
\newblock \emph{Intelligent systems: A modern approach}, pages 149--185.

\bibitem[{Gu et~al.(2024)Gu, Jiang, Shi, Tan, Zhai, Xu, Li, Shen, Ma, Liu et~al.}]{gu2024survey}
Jiawei Gu, Xuhui Jiang, Zhichao Shi, Hexiang Tan, Xuehao Zhai, Chengjin Xu, Wei Li, Yinghan Shen, Shengjie Ma, Honghao Liu, and 1 others. 2024.
\newblock A survey on llm-as-a-judge.
\newblock \emph{arXiv preprint arXiv:2411.15594}.

\bibitem[{Han et~al.(2023{\natexlab{a}})Han, Yoo, Kim, Myung, Kim, Lim, Kim, Lee, Hong, Ahn et~al.}]{han2023recipe}
Jieun Han, Haneul Yoo, Yoonsu Kim, Junho Myung, Minsun Kim, Hyunseung Lim, Juho Kim, Tak~Yeon Lee, Hwajung Hong, So-Yeon Ahn, and 1 others. 2023{\natexlab{a}}.
\newblock Recipe: How to integrate chatgpt into efl writing education.
\newblock In \emph{Proceedings of the tenth ACM conference on learning@ scale}, pages 416--420.

\bibitem[{Han et~al.(2024)Han, Yoo, Myung, Kim, Lim, Kim, Lee, Hong, Kim, Ahn, and Oh}]{han-etal-2024-llm}
Jieun Han, Haneul Yoo, Junho Myung, Minsun Kim, Hyunseung Lim, Yoonsu Kim, Tak~Yeon Lee, Hwajung Hong, Juho Kim, So-Yeon Ahn, and Alice Oh. 2024.
\newblock \href {https://doi.org/10.18653/v1/2024.customnlp4u-1.21} {{LLM}-as-a-tutor in {EFL} writing education: Focusing on evaluation of student-{LLM} interaction}.
\newblock In \emph{Proceedings of the 1st Workshop on Customizable NLP: Progress and Challenges in Customizing NLP for a Domain, Application, Group, or Individual (CustomNLP4U)}, pages 284--293, Miami, Florida, USA. Association for Computational Linguistics.

\bibitem[{Han et~al.(2023{\natexlab{b}})Han, Yoo, Myung, Kim, Lim, Kim, Lee, Hong, Kim, Ahn et~al.}]{han2023fabric}
Jieun Han, Haneul Yoo, Junho Myung, Minsun Kim, Hyunseung Lim, Yoonsu Kim, Tak~Yeon Lee, Hwajung Hong, Juho Kim, So-Yeon Ahn, and 1 others. 2023{\natexlab{b}}.
\newblock Fabric: Automated scoring and feedback generation for essays.
\newblock \emph{arXiv preprint arXiv:2310.05191}.

\bibitem[{Hart(1981)}]{hart1981language}
Robert Hart. 1981.
\newblock Language study and the plato system.
\newblock \emph{Studies in language learning}, 3(1):1--24.

\bibitem[{He and Li(2024)}]{he2024mitigating}
Lin He and Keqin Li. 2024.
\newblock Mitigating hallucinations in llm using k-means clustering of synonym semantic relevance.
\newblock \emph{Authorea Preprints}.

\bibitem[{Henkel et~al.(2024)Henkel, Hills, Boxer, Roberts, and Levonian}]{henkel2024can}
Owen Henkel, Libby Hills, Adam Boxer, Bill Roberts, and Zach Levonian. 2024.
\newblock Can large language models make the grade? an empirical study evaluating llms ability to mark short answer questions in k-12 education.
\newblock In \emph{Proceedings of the Eleventh ACM Conference on Learning@ Scale}, pages 300--304.

\bibitem[{Ho et~al.(2024)Ho, Ly, and Nguyen}]{ho2024mitigating}
Huu-Tuong Ho, Duc-Tin Ly, and Luong~Vuong Nguyen. 2024.
\newblock Mitigating hallucinations in large language models for educational application.
\newblock In \emph{2024 IEEE International Conference on Consumer Electronics-Asia (ICCE-Asia)}, pages 1--4. IEEE.

\bibitem[{Hou(2020)}]{hou2020foreign}
Yanxia Hou. 2020.
\newblock Foreign language education in the era of artificial intelligence.
\newblock In \emph{Big Data Analytics for Cyber-Physical System in Smart City: BDCPS 2019, 28-29 December 2019, Shenyang, China}, pages 937--944. Springer.

\bibitem[{Hu et~al.(2024)Hu, Zheng, Zhu, Ding, Wang, and Gu}]{hu2024teaching}
Bihao Hu, Longwei Zheng, Jiayi Zhu, Lishan Ding, Yilei Wang, and Xiaoqing Gu. 2024.
\newblock Teaching plan generation and evaluation with gpt-4: Unleashing the potential of llm in instructional design.
\newblock \emph{IEEE Transactions on Learning Technologies}.

\bibitem[{Hu and Wang(2024)}]{hu2024foke}
Silan Hu and Xiaoning Wang. 2024.
\newblock Foke: A personalized and explainable education framework integrating foundation models, knowledge graphs, and prompt engineering.
\newblock In \emph{China National Conference on Big Data and Social Computing}, pages 399--411. Springer.

\bibitem[{Huang et~al.(2024{\natexlab{a}})Huang, Wei, and Huang}]{huang2024generating}
Chieh-Yang Huang, Jing Wei, and Ting-Hao~Kenneth Huang. 2024{\natexlab{a}}.
\newblock Generating educational materials with different levels of readability using llms.
\newblock In \emph{Proceedings of the Third Workshop on Intelligent and Interactive Writing Assistants}, pages 16--22.

\bibitem[{Huang et~al.(2023)Huang, Yu, Ma, Zhong, Feng, Wang, Chen, Peng, Feng, Qin et~al.}]{huang2023survey}
Lei Huang, Weijiang Yu, Weitao Ma, Weihong Zhong, Zhangyin Feng, Haotian Wang, Qianglong Chen, Weihua Peng, Xiaocheng Feng, Bing Qin, and 1 others. 2023.
\newblock A survey on hallucination in large language models: Principles, taxonomy, challenges, and open questions.
\newblock \emph{ACM Transactions on Information Systems}.

\bibitem[{Huang et~al.(2024{\natexlab{b}})Huang, Liu, Chen, Wang, Wang, Lian, Wang, Tang, and Chen}]{huang2024understanding}
Xu~Huang, Weiwen Liu, Xiaolong Chen, Xingmei Wang, Hao Wang, Defu Lian, Yasheng Wang, Ruiming Tang, and Enhong Chen. 2024{\natexlab{b}}.
\newblock Understanding the planning of llm agents: A survey.
\newblock \emph{arXiv preprint arXiv:2402.02716}.

\bibitem[{Hyland and Hyland(2006)}]{hyland2006feedback}
Ken Hyland and Fiona Hyland. 2006.
\newblock Feedback on second language students' writing.
\newblock \emph{Language teaching}, 39(2):83--101.

\bibitem[{Imperial et~al.(2024)Imperial, Forey, and Tayyar~Madabushi}]{imperial-etal-2024-standardize}
Joseph~Marvin Imperial, Gail Forey, and Harish Tayyar~Madabushi. 2024.
\newblock \href {https://doi.org/10.18653/v1/2024.emnlp-main.94} {Standardize: Aligning language models with expert-defined standards for content generation}.
\newblock In \emph{Proceedings of the 2024 Conference on Empirical Methods in Natural Language Processing}, pages 1573--1594, Miami, Florida, USA. Association for Computational Linguistics.

\bibitem[{Jadhav et~al.(2024)Jadhav, Shanbhag, Thakurdesai, Sinare, and Joshi}]{jadhav2024limitations}
Suramya Jadhav, Abhay Shanbhag, Amogh Thakurdesai, Ridhima Sinare, and Raviraj Joshi. 2024.
\newblock On limitations of llm as annotator for low resource languages.
\newblock \emph{arXiv preprint arXiv:2411.17637}.

\bibitem[{James(2013)}]{james2013errors}
Carl James. 2013.
\newblock \emph{Errors in language learning and use: Exploring error analysis}.
\newblock Routledge.

\bibitem[{Jeon and Lee(2023)}]{jeon2023large}
Jaeho Jeon and Seongyong Lee. 2023.
\newblock Large language models in education: A focus on the complementary relationship between human teachers and chatgpt.
\newblock \emph{Education and Information Technologies}, 28(12):15873--15892.

\bibitem[{Jiang et~al.(2024{\natexlab{a}})Jiang, Li, Zhou, Qi, Hu, Wei, Jiang, and Wu}]{jiang2024ai}
Yuan-Hao Jiang, Ruijia Li, Yizhou Zhou, Changyong Qi, Hanglei Hu, Yuang Wei, Bo~Jiang, and Yonghe Wu. 2024{\natexlab{a}}.
\newblock Ai agent for education: von neumann multi-agent system framework.
\newblock \emph{arXiv preprint arXiv:2501.00083}.

\bibitem[{Jiang et~al.(2024{\natexlab{b}})Jiang, Peng, Feng, Li, and Li}]{jiang2024llms}
Zhuoxuan Jiang, Haoyuan Peng, Shanshan Feng, Fan Li, and Dongsheng Li. 2024{\natexlab{b}}.
\newblock Llms can find mathematical reasoning mistakes by pedagogical chain-of-thought.
\newblock \emph{arXiv preprint arXiv:2405.06705}.

\bibitem[{Jung et~al.(2024)Jung, Yoo, Yoon, and Jang}]{jung2024clst}
Heeseok Jung, Jaesang Yoo, Yohaan Yoon, and Yeonju Jang. 2024.
\newblock Clst: Cold-start mitigation in knowledge tracing by aligning a generative language model as a students' knowledge tracer.
\newblock \emph{arXiv preprint arXiv:2406.10296}.

\bibitem[{Kamoi et~al.(2024)Kamoi, Das, Lou, Ahn, Zhao, Lu, Zhang, Zhang, Zhang, Vummanthala, Dave, Qin, Cohan, Yin, and Zhang}]{kamoi2024evaluating}
Ryo Kamoi, Sarkar Snigdha~Sarathi Das, Renze Lou, Jihyun~Janice Ahn, Yilun Zhao, Xiaoxin Lu, Nan Zhang, Yusen Zhang, Haoran~Ranran Zhang, Sujeeth~Reddy Vummanthala, Salika Dave, Shaobo Qin, Arman Cohan, Wenpeng Yin, and Rui Zhang. 2024.
\newblock \href {https://openreview.net/forum?id=dnwRScljXr} {Evaluating {LLM}s at detecting errors in {LLM} responses}.
\newblock In \emph{First Conference on Language Modeling}.

\bibitem[{Karata{\c{s}} et~al.(2024)Karata{\c{s}}, Abedi, Ozek~Gunyel, Karadeniz, and Kuzgun}]{karatacs2024incorporating}
Fatih Karata{\c{s}}, Faramarz~Ya{\c{s}}ar Abedi, Filiz Ozek~Gunyel, Derya Karadeniz, and Yasemin Kuzgun. 2024.
\newblock Incorporating ai in foreign language education: An investigation into chatgpt’s effect on foreign language learners.
\newblock \emph{Education and Information Technologies}, pages 1--24.

\bibitem[{K{\"a}ser and Alexandron(2024)}]{kaser2024simulated}
Tanja K{\"a}ser and Giora Alexandron. 2024.
\newblock Simulated learners in educational technology: A systematic literature review and a turing-like test.
\newblock \emph{International Journal of Artificial Intelligence in Education}, 34(2):545--585.

\bibitem[{Katinskaia(2025)}]{katinskaia2025overview}
Anisia Katinskaia. 2025.
\newblock An overview of artificial intelligence in computer-assisted language learning.
\newblock \emph{arXiv preprint arXiv:2505.02032}.

\bibitem[{Katuka et~al.(2024)Katuka, Gain, and Yu}]{katuka2024investigating}
Gloria~Ashiya Katuka, Alexander Gain, and Yen-Yun Yu. 2024.
\newblock Investigating automatic scoring and feedback using large language models.
\newblock \emph{arXiv preprint arXiv:2405.00602}.

\bibitem[{Kharchenko et~al.(2024)Kharchenko, Roosta, Chadha, and Shah}]{kharchenko2024well}
Julia Kharchenko, Tanya Roosta, Aman Chadha, and Chirag Shah. 2024.
\newblock How well do llms represent values across cultures? empirical analysis of llm responses based on hofstede cultural dimensions.
\newblock \emph{arXiv preprint arXiv:2406.14805}.

\bibitem[{Kim et~al.(2022)Kim, Kim, Yoo, and Kang}]{kim-etal-2022-generating}
Gangwoo Kim, Sungdong Kim, Kang~Min Yoo, and Jaewoo Kang. 2022.
\newblock \href {https://doi.org/10.18653/v1/2022.emnlp-main.151} {Generating information-seeking conversations from unlabeled documents}.
\newblock In \emph{Proceedings of the 2022 Conference on Empirical Methods in Natural Language Processing}, pages 2362--2378, Abu Dhabi, United Arab Emirates. Association for Computational Linguistics.

\bibitem[{Kim et~al.(2024)Kim, Mitra, Li~Chen, Rahman, and Zhang}]{kim-etal-2024-meganno}
Hannah Kim, Kushan Mitra, Rafael Li~Chen, Sajjadur Rahman, and Dan Zhang. 2024.
\newblock \href {https://aclanthology.org/2024.eacl-demo.18/} {{MEGA}nno+: A human-{LLM} collaborative annotation system}.
\newblock In \emph{Proceedings of the 18th Conference of the European Chapter of the Association for Computational Linguistics: System Demonstrations}, pages 168--176, St. Julians, Malta. Association for Computational Linguistics.

\bibitem[{KIM et~al.(2025)KIM, Park, Jeon, Fels, Dodson, and Seo}]{kim2025augmented}
HYEJI KIM, Jongyoul Park, Hyeongbae Jeon, Sidney~S Fels, Samuel Dodson, and Kyoungwon Seo. 2025.
\newblock Augmented educators and ai: Shaping the future of human-ai collaboration in learning.
\newblock In \emph{Proceedings of the Extended Abstracts of the CHI Conference on Human Factors in Computing Systems}, pages 1--6.

\bibitem[{Kundu and Barbosa(2024)}]{kundu2024large}
Anindita Kundu and Denilson Barbosa. 2024.
\newblock Are large language models good essay graders?
\newblock \emph{arXiv preprint arXiv:2409.13120}.

\bibitem[{Kurdi et~al.(2020)Kurdi, Leo, Parsia, Sattler, and Al-Emari}]{kurdi2020systematic}
Ghader Kurdi, Jared Leo, Bijan Parsia, Uli Sattler, and Salam Al-Emari. 2020.
\newblock A systematic review of automatic question generation for educational purposes.
\newblock \emph{International Journal of Artificial Intelligence in Education}, 30:121--204.

\bibitem[{Kwon et~al.(2024)Kwon, Kim, Park, Lee, and Kim}]{kwon-etal-2024-biped}
Soonwoo Kwon, Sojung Kim, Minju Park, Seunghyun Lee, and Kyuseok Kim. 2024.
\newblock \href {https://doi.org/10.18653/v1/2024.acl-long.186} {{BIPED}: Pedagogically informed tutoring system for {ESL} education}.
\newblock In \emph{Proceedings of the 62nd Annual Meeting of the Association for Computational Linguistics (Volume 1: Long Papers)}, pages 3389--3414, Bangkok, Thailand. Association for Computational Linguistics.

\bibitem[{Lee et~al.(2024{\natexlab{a}})Lee, Huang, Cho, Roh, Kwon, and Lee}]{lee2024developing}
Jeongmin Lee, Jin-Xia Huang, Minsoo Cho, Yoon-Hyung Roh, Oh-Woog Kwon, and Yunkeun Lee. 2024{\natexlab{a}}.
\newblock Developing conversational intelligent tutoring for speaking skills in second language learning.
\newblock In \emph{International Conference on Intelligent Tutoring Systems}, pages 131--148. Springer.

\bibitem[{Lee et~al.(2024{\natexlab{b}})Lee, Kim, Khetan, and Kang}]{lee2024human}
Minhwa Lee, Zae~Myung Kim, Vivek Khetan, and Dongyeop Kang. 2024{\natexlab{b}}.
\newblock Human-ai collaborative taxonomy construction: A case study in profession-specific writing assistants.
\newblock In \emph{Proceedings of the Third Workshop on Intelligent and Interactive Writing Assistants}, pages 51--57.

\bibitem[{Lee et~al.(2024{\natexlab{c}})Lee, Jung, Jeon, Sohn, Hwang, Moon, and Kim}]{lee2024few}
Unggi Lee, Haewon Jung, Younghoon Jeon, Younghoon Sohn, Wonhee Hwang, Jewoong Moon, and Hyeoncheol Kim. 2024{\natexlab{c}}.
\newblock Few-shot is enough: exploring chatgpt prompt engineering method for automatic question generation in english education.
\newblock \emph{Education and Information Technologies}, 29(9):11483--11515.

\bibitem[{Leu~Jr(1982)}]{leu1982oral}
Donald~J Leu~Jr. 1982.
\newblock Oral reading error analysis: A critical review of research and application.
\newblock \emph{Reading Research Quarterly}, pages 420--437.

\bibitem[{Li et~al.(2024{\natexlab{a}})Li, Jiang, Huang, Beigi, Zhao, Tan, Bhattacharjee, Jiang, Chen, Wu et~al.}]{li2024generation}
Dawei Li, Bohan Jiang, Liangjie Huang, Alimohammad Beigi, Chengshuai Zhao, Zhen Tan, Amrita Bhattacharjee, Yuxuan Jiang, Canyu Chen, Tianhao Wu, and 1 others. 2024{\natexlab{a}}.
\newblock From generation to judgment: Opportunities and challenges of llm-as-a-judge.
\newblock \emph{arXiv preprint arXiv:2411.16594}.

\bibitem[{Li et~al.(2024{\natexlab{b}})Li, Dong, Chen, Su, Zhou, Ai, Ye, and Liu}]{li2024llms}
Haitao Li, Qian Dong, Junjie Chen, Huixue Su, Yujia Zhou, Qingyao Ai, Ziyi Ye, and Yiqun Liu. 2024{\natexlab{b}}.
\newblock Llms-as-judges: A comprehensive survey on llm-based evaluation methods.
\newblock \emph{arXiv preprint arXiv:2412.05579}.

\bibitem[{Li and Zhang(2024)}]{li-zhang-2024-planning}
Kunze Li and Yu~Zhang. 2024.
\newblock \href {https://doi.org/10.18653/v1/2024.findings-acl.280} {Planning first, question second: An {LLM}-guided method for controllable question generation}.
\newblock In \emph{Findings of the Association for Computational Linguistics: ACL 2024}, pages 4715--4729, Bangkok, Thailand. Association for Computational Linguistics.

\bibitem[{Li et~al.(2023)Li, Shi, Ziems, Kan, Chen, Liu, and Yang}]{li-etal-2023-coannotating}
Minzhi Li, Taiwei Shi, Caleb Ziems, Min-Yen Kan, Nancy Chen, Zhengyuan Liu, and Diyi Yang. 2023.
\newblock \href {https://doi.org/10.18653/v1/2023.emnlp-main.92} {{C}o{A}nnotating: Uncertainty-guided work allocation between human and large language models for data annotation}.
\newblock In \emph{Proceedings of the 2023 Conference on Empirical Methods in Natural Language Processing}, pages 1487--1505, Singapore. Association for Computational Linguistics.

\bibitem[{Li and Liu(2024)}]{li2024applying}
Wenchao Li and Haitao Liu. 2024.
\newblock Applying large language models for automated essay scoring for non-native japanese.
\newblock \emph{Humanities and Social Sciences Communications}, 11(1):1--15.

\bibitem[{Li et~al.(2024{\natexlab{c}})Li, Henriksson, Duneld, Nouri, and Wu}]{li2024supporting}
Xiu Li, Aron Henriksson, Martin Duneld, Jalal Nouri, and Yongchao Wu. 2024{\natexlab{c}}.
\newblock Supporting teaching-to-the-curriculum by linking diagnostic tests to curriculum goals: Using textbook content as context for retrieval-augmented generation with large language models.
\newblock In \emph{International Conference on Artificial Intelligence in Education}, pages 118--132. Springer.

\bibitem[{Liang et~al.(2024)Liang, Wang, Wang, Song, Yang, Niu, Hu, Liu, Yao, Xiong et~al.}]{liang2024controllable}
Xun Liang, Hanyu Wang, Yezhaohui Wang, Shichao Song, Jiawei Yang, Simin Niu, Jie Hu, Dan Liu, Shunyu Yao, Feiyu Xiong, and 1 others. 2024.
\newblock Controllable text generation for large language models: A survey.
\newblock \emph{arXiv preprint arXiv:2408.12599}.

\bibitem[{Liu et~al.(2024{\natexlab{a}})Liu, Gurevych, and Korhonen}]{liu2024culturally}
Chen~Cecilia Liu, Iryna Gurevych, and Anna Korhonen. 2024{\natexlab{a}}.
\newblock Culturally aware and adapted nlp: A taxonomy and a survey of the state of the art.
\newblock \emph{arXiv preprint arXiv:2406.03930}.

\bibitem[{Liu et~al.(2024{\natexlab{b}})Liu, Huang, Xiao, Sha, Wu, Liu, Wang, and Chen}]{liusocraticlm}
Jiayu Liu, Zhenya Huang, Tong Xiao, Jing Sha, Jinze Wu, Qi~Liu, Shijin Wang, and Enhong Chen. 2024{\natexlab{b}}.
\newblock Socraticlm: Exploring socratic personalized teaching with large language models.
\newblock In \emph{The Thirty-eighth Annual Conference on Neural Information Processing Systems}.

\bibitem[{Liu et~al.(2024{\natexlab{c}})Liu, Wei, Liu, Si, Zhang, Rao, Zheng, Peng, Yang, Zhou et~al.}]{liu2024best}
Ruibo Liu, Jerry Wei, Fangyu Liu, Chenglei Si, Yanzhe Zhang, Jinmeng Rao, Steven Zheng, Daiyi Peng, Diyi Yang, Denny Zhou, and 1 others. 2024{\natexlab{c}}.
\newblock Best practices and lessons learned on synthetic data for language models.
\newblock \emph{arXiv preprint arXiv:2404.07503}.

\bibitem[{Liu et~al.(2024{\natexlab{d}})Liu, Yin, Lin, and Chen}]{liu2024personality}
Zhengyuan Liu, Stella~Xin Yin, Geyu Lin, and Nancy~F Chen. 2024{\natexlab{d}}.
\newblock Personality-aware student simulation for conversational intelligent tutoring systems.
\newblock \emph{arXiv preprint arXiv:2404.06762}.

\bibitem[{Liu et~al.(2025)Liu, Litman, Wang, Li, Gobat, Matsumura, and Correnti}]{liu2025erevise+}
Zhexiong Liu, Diane Litman, Elaine Wang, Tianwen Li, Mason Gobat, Lindsay~Clare Matsumura, and Richard Correnti. 2025.
\newblock erevise+ rf: A writing evaluation system for assessing student essay revisions and providing formative feedback.
\newblock \emph{arXiv preprint arXiv:2501.00715}.

\bibitem[{Long et~al.(2024)Long, Wang, Xiao, Zhao, Ding, Chen, and Wang}]{long-etal-2024-llms}
Lin Long, Rui Wang, Ruixuan Xiao, Junbo Zhao, Xiao Ding, Gang Chen, and Haobo Wang. 2024.
\newblock \href {https://doi.org/10.18653/v1/2024.findings-acl.658} {On {LLM}s-driven synthetic data generation, curation, and evaluation: A survey}.
\newblock In \emph{Findings of the Association for Computational Linguistics: ACL 2024}, pages 11065--11082, Bangkok, Thailand. Association for Computational Linguistics.

\bibitem[{Lu et~al.(2024)Lu, Qiu, Ding, Zhang, Kocmi, and Tao}]{lu2023error}
Qingyu Lu, Baopu Qiu, Liang Ding, Kanjian Zhang, Tom Kocmi, and Dacheng Tao. 2024.
\newblock \href {https://doi.org/10.18653/v1/2024.findings-acl.520} {Error analysis prompting enables human-like translation evaluation in large language models}.
\newblock In \emph{Findings of the Association for Computational Linguistics: ACL 2024}, pages 8801--8816, Bangkok, Thailand. Association for Computational Linguistics.

\bibitem[{Lu and Wang(2024)}]{lu2024generative}
Xinyi Lu and Xu~Wang. 2024.
\newblock Generative students: Using llm-simulated student profiles to support question item evaluation.
\newblock \emph{arXiv preprint arXiv:2405.11591}.

\bibitem[{Ma and Guo(2019)}]{ma2019cognitive}
Wenchao Ma and Wenjing Guo. 2019.
\newblock Cognitive diagnosis models for multiple strategies.
\newblock \emph{British Journal of Mathematical and Statistical Psychology}, 72(2):370--392.

\bibitem[{Macina et~al.(2025)Macina, Daheim, Hakimi, Kapur, Gurevych, and Sachan}]{macina2025mathtutorbench}
Jakub Macina, Nico Daheim, Ido Hakimi, Manu Kapur, Iryna Gurevych, and Mrinmaya Sachan. 2025.
\newblock Mathtutorbench: A benchmark for measuring open-ended pedagogical capabilities of llm tutors.
\newblock \emph{arXiv preprint arXiv:2502.18940}.

\bibitem[{Mashoor and Abdullah(2020)}]{mashoor2020error}
Bakheet Bayan~Nayif Mashoor and ATH~bin Abdullah. 2020.
\newblock Error analysis of spoken english language among jordanian secondary school students.
\newblock \emph{International Journal of Education and Research}, 8(5):75--82.

\bibitem[{Maurya et~al.(2024)Maurya, Srivatsa, Petukhova, and Kochmar}]{maurya2024unifying}
Kaushal~Kumar Maurya, KV~Srivatsa, Kseniia Petukhova, and Ekaterina Kochmar. 2024.
\newblock Unifying ai tutor evaluation: An evaluation taxonomy for pedagogical ability assessment of llm-powered ai tutors.
\newblock \emph{arXiv preprint arXiv:2412.09416}.

\bibitem[{Mizumoto and Eguchi(2023)}]{mizumoto2023exploring}
Atsushi Mizumoto and Masaki Eguchi. 2023.
\newblock Exploring the potential of using an ai language model for automated essay scoring.
\newblock \emph{Research Methods in Applied Linguistics}, 2(2):100050.

\bibitem[{Modran et~al.(2024)Modran, Bogdan, Ursuțiu, Samoila, and Modran}]{modran2024llm}
Horia Modran, Ioana~Corina Bogdan, Doru Ursuțiu, Cornel Samoila, and Paul~Livius Modran. 2024.
\newblock Llm intelligent agent tutoring in higher education courses using a rag approach.
\newblock \emph{Preprints 2024}, 2024070519.

\bibitem[{Mulla and Gharpure(2023)}]{mulla2023automatic}
Nikahat Mulla and Prachi Gharpure. 2023.
\newblock Automatic question generation: a review of methodologies, datasets, evaluation metrics, and applications.
\newblock \emph{Progress in Artificial Intelligence}, 12(1):1--32.

\bibitem[{Myles(2002)}]{myles2002second}
Johanne Myles. 2002.
\newblock Second language writing and research: The writing process and error analysis in student texts.
\newblock \emph{Tesl-ej}, 6(2):1--20.

\bibitem[{Nahar et~al.(2024)Nahar, Seo, Lee, Xiong, and Lee}]{nahar2024fakes}
Mahjabin Nahar, Haeseung Seo, Eun-Ju Lee, Aiping Xiong, and Dongwon Lee. 2024.
\newblock \href {https://openreview.net/forum?id=c30qeMg8dv} {Fakes of varying shades: How warning affects human perception and engagement regarding {LLM} hallucinations}.
\newblock In \emph{First Conference on Language Modeling}.

\bibitem[{Neshaei et~al.(2024)Neshaei, Davis, Hazimeh, Lazarevski, Dillenbourg, and K{\"a}ser}]{neshaei2024towards}
Seyed~Parsa Neshaei, Richard~Lee Davis, Adam Hazimeh, Bojan Lazarevski, Pierre Dillenbourg, and Tanja K{\"a}ser. 2024.
\newblock Towards modeling learner performance with large language models.
\newblock \emph{arXiv preprint arXiv:2403.14661}.

\bibitem[{Nicholls et~al.(2024)Nicholls, Caines, and Buttery}]{nicholls2024write}
Diane Nicholls, Andrew Caines, and Paula Buttery. 2024.
\newblock The write \& improve corpus 2024: Error-annotated and cefr-labelled essays by learners of english.

\bibitem[{Nunan(1989)}]{nunan1989designing}
David Nunan. 1989.
\newblock \emph{Designing tasks for the communicative classroom}.
\newblock Cambridge university press.

\bibitem[{Och(2006)}]{och2006statistical}
Franz Och. 2006.
\newblock \href {https://research.google/blog/statistical-machine-translation-live/} {Statistical machine translation live}.

\bibitem[{Pal~Chowdhury et~al.(2024)Pal~Chowdhury, Zouhar, and Sachan}]{pal2024autotutor}
Sankalan Pal~Chowdhury, Vil{\'e}m Zouhar, and Mrinmaya Sachan. 2024.
\newblock Autotutor meets large language models: A language model tutor with rich pedagogy and guardrails.
\newblock In \emph{Proceedings of the Eleventh ACM Conference on Learning@ Scale}, pages 5--15.

\bibitem[{Paul and Elder(2007)}]{paul2007critical}
Richard Paul and Linda Elder. 2007.
\newblock Critical thinking: The art of socratic questioning.
\newblock \emph{Journal of developmental education}, 31(1):36.

\bibitem[{Radford et~al.(2009)Radford, Atkinson, Britain, Clahsen, and Spencer}]{radford2009linguistics}
Andrew Radford, Martin Atkinson, David Britain, Harald Clahsen, and Andrew Spencer. 2009.
\newblock \emph{Linguistics: an introduction}.
\newblock Cambridge University Press.

\bibitem[{Rafailov et~al.(2024)Rafailov, Sharma, Mitchell, Manning, Ermon, and Finn}]{rafailov2024direct}
Rafael Rafailov, Archit Sharma, Eric Mitchell, Christopher~D Manning, Stefano Ermon, and Chelsea Finn. 2024.
\newblock Direct preference optimization: Your language model is secretly a reward model.
\newblock \emph{Advances in Neural Information Processing Systems}, 36.

\bibitem[{Rashov(2024)}]{rashov2024modern}
Oybek Rashov. 2024.
\newblock Modern methods of teaching foreign languages.
\newblock In \emph{International Scientific and Current Research Conferences}, pages 158--164.

\bibitem[{Rathod et~al.(2022)Rathod, Tu, and Stasaski}]{rathod-etal-2022-educational}
Manav Rathod, Tony Tu, and Katherine Stasaski. 2022.
\newblock \href {https://doi.org/10.18653/v1/2022.bea-1.26} {Educational multi-question generation for reading comprehension}.
\newblock In \emph{Proceedings of the 17th Workshop on Innovative Use of NLP for Building Educational Applications (BEA 2022)}, pages 216--223, Seattle, Washington. Association for Computational Linguistics.

\bibitem[{Razafinirina et~al.(2024)Razafinirina, Dimbisoa, and Mahatody}]{razafinirina2024pedagogical}
Mahefa~Abel Razafinirina, William~Germain Dimbisoa, and Thomas Mahatody. 2024.
\newblock Pedagogical alignment of large language models (llm) for personalized learning: A survey, trends and challenges.
\newblock \emph{Journal of Intelligent Learning Systems and Applications}, 16(4):448--480.

\bibitem[{Samuel et~al.(2024)Samuel, Aynaou, Chowdhury, Venkat~Ramanan, and Chadha}]{samuel-etal-2024-llms}
Vinay Samuel, Houda Aynaou, Arijit Chowdhury, Karthik Venkat~Ramanan, and Aman Chadha. 2024.
\newblock \href {https://doi.org/10.18653/v1/2024.acl-srw.36} {Can {LLM}s augment low-resource reading comprehension datasets? opportunities and challenges}.
\newblock In \emph{Proceedings of the 62nd Annual Meeting of the Association for Computational Linguistics (Volume 4: Student Research Workshop)}, pages 307--317, Bangkok, Thailand. Association for Computational Linguistics.

\bibitem[{Scarlatos and Lan(2024)}]{scarlatos2024exploring}
Alexander Scarlatos and Andrew Lan. 2024.
\newblock Exploring knowledge tracing in tutor-student dialogues.
\newblock \emph{arXiv preprint arXiv:2409.16490}.

\bibitem[{Scarlatos et~al.(2025)Scarlatos, Liu, Lee, Baraniuk, and Lan}]{scarlatos2025training}
Alexander Scarlatos, Naiming Liu, Jaewook Lee, Richard Baraniuk, and Andrew Lan. 2025.
\newblock Training llm-based tutors to improve student learning outcomes in dialogues.
\newblock \emph{arXiv preprint arXiv:2503.06424}.

\bibitem[{Schmidt and Strasser(2022)}]{schmidt2022artificial}
Torben Schmidt and Thomas Strasser. 2022.
\newblock Artificial intelligence in foreign language learning and teaching: a call for intelligent practice.
\newblock \emph{Anglistik: International Journal of English Studies}, 33(1):165--184.

\bibitem[{Schmucker et~al.(2024)Schmucker, Xia, Azaria, and Mitchell}]{schmucker2024ruffle}
Robin Schmucker, Meng Xia, Amos Azaria, and Tom Mitchell. 2024.
\newblock Ruffle\&riley: Insights from designing and evaluating a large language model-based conversational tutoring system.
\newblock In \emph{International Conference on Artificial Intelligence in Education}, pages 75--90. Springer.

\bibitem[{Schneider et~al.(2023)Schneider, Schenk, and Niklaus}]{schneider2023towards}
Johannes Schneider, Bernd Schenk, and Christina Niklaus. 2023.
\newblock Towards llm-based autograding for short textual answers.
\newblock \emph{arXiv preprint arXiv:2309.11508}.

\bibitem[{Se{\ss}ler et~al.(2024)Se{\ss}ler, F{\"u}rstenberg, B{\"u}hler, and Kasneci}]{sessler2024can}
Kathrin Se{\ss}ler, Maurice F{\"u}rstenberg, Babette B{\"u}hler, and Enkelejda Kasneci. 2024.
\newblock Can ai grade your essays? a comparative analysis of large language models and teacher ratings in multidimensional essay scoring.
\newblock \emph{arXiv preprint arXiv:2411.16337}.

\bibitem[{Shao et~al.(2025)Shao, Yuan, Gao, He, Yang, and Chen}]{shao2025unlocking}
Zekai Shao, Siyu Yuan, Lin Gao, Yixuan He, Deqing Yang, and Siming Chen. 2025.
\newblock Unlocking scientific concepts: How effective are llm-generated analogies for student understanding and classroom practice?
\newblock \emph{arXiv preprint arXiv:2502.16895}.

\bibitem[{Sharma et~al.(2024)Sharma, Mhasakar, Mehra, Venaik, Singhal, Kumar, and Mittal}]{sharma2024comuniqa}
Shikhar Sharma, Manas Mhasakar, Apurv Mehra, Utkarsh Venaik, Ujjwal Singhal, Dhruv Kumar, and Kashish Mittal. 2024.
\newblock Comuniqa: Exploring large language models for improving english speaking skills.
\newblock In \emph{Proceedings of the 7th ACM SIGCAS/SIGCHI Conference on Computing and Sustainable Societies}, pages 256--267.

\bibitem[{Shen et~al.(2024{\natexlab{a}})Shen, Tan, Chen, Chen, Zhang, Xu, Zheng, Koehn, and Khashabi}]{shen2024language}
Lingfeng Shen, Weiting Tan, Sihao Chen, Yunmo Chen, Jingyu Zhang, Haoran Xu, Boyuan Zheng, Philipp Koehn, and Daniel Khashabi. 2024{\natexlab{a}}.
\newblock The language barrier: Dissecting safety challenges of llms in multilingual contexts.
\newblock \emph{arXiv preprint arXiv:2401.13136}.

\bibitem[{Shen et~al.(2024{\natexlab{b}})Shen, Liu, Huang, Zheng, Yin, Wang, and Chen}]{shen2024survey}
Shuanghong Shen, Qi~Liu, Zhenya Huang, Yonghe Zheng, Minghao Yin, Minjuan Wang, and Enhong Chen. 2024{\natexlab{b}}.
\newblock A survey of knowledge tracing: Models, variants, and applications.
\newblock \emph{IEEE Transactions on Learning Technologies}.

\bibitem[{Shi et~al.(2025)Shi, Liang, and Xu}]{shi2025educationq}
Yao Shi, Rongkeng Liang, and Yong Xu. 2025.
\newblock Educationq: Evaluating llms' teaching capabilities through multi-agent dialogue framework.
\newblock \emph{arXiv preprint arXiv:2504.14928}.

\bibitem[{Shojaei et~al.(2025)Shojaei, Gulati, Jasperson, Wang, Cimolato, Cao, Neiswanger, and Garikipati}]{shojaei2025ai}
Mostafa~Faghih Shojaei, Rahul Gulati, Benjamin~A Jasperson, Shangshang Wang, Simone Cimolato, Dangli Cao, Willie Neiswanger, and Krishna Garikipati. 2025.
\newblock Ai-university: An llm-based platform for instructional alignment to scientific classrooms.
\newblock \emph{arXiv preprint arXiv:2504.08846}.

\bibitem[{Shorten et~al.(2021)Shorten, Khoshgoftaar, and Furht}]{shorten2021text}
Connor Shorten, Taghi~M Khoshgoftaar, and Borko Furht. 2021.
\newblock Text data augmentation for deep learning.
\newblock \emph{Journal of big Data}, 8(1):101.

\bibitem[{Siyan et~al.(2024)Siyan, Shao, Yu, and Hirschberg}]{li-etal-2024-eden}
Li~Siyan, Teresa Shao, Zhou Yu, and Julia Hirschberg. 2024.
\newblock \href {https://doi.org/10.18653/v1/2024.findings-emnlp.200} {{EDEN}: Empathetic dialogues for {E}nglish learning}.
\newblock In \emph{Findings of the Association for Computational Linguistics: EMNLP 2024}, pages 3492--3511, Miami, Florida, USA. Association for Computational Linguistics.

\bibitem[{Song et~al.(2024{\natexlab{a}})Song, Huang, Zhou, and Ma}]{song2024multilingual}
Jiayang Song, Yuheng Huang, Zhehua Zhou, and Lei Ma. 2024{\natexlab{a}}.
\newblock Multilingual blending: Llm safety alignment evaluation with language mixture.
\newblock \emph{arXiv preprint arXiv:2407.07342}.

\bibitem[{Song et~al.(2025)Song, Yuk, Choi, Yoo, Lim, Lim, and Park}]{song-etal-2025-unified}
SeungWoo Song, Junghun Yuk, ChangSu Choi, HanGyeol Yoo, HyeonSeok Lim, KyungTae Lim, and Jungyeul Park. 2025.
\newblock \href {https://aclanthology.org/2025.findings-naacl.250/} {Unified automated essay scoring and grammatical error correction}.
\newblock In \emph{Findings of the Association for Computational Linguistics: NAACL 2025}, pages 4412--4426, Albuquerque, New Mexico. Association for Computational Linguistics.

\bibitem[{Song et~al.(2024{\natexlab{b}})Song, Krishna, Bhatt, Gimpel, and Iyyer}]{song-etal-2024-gee}
Yixiao Song, Kalpesh Krishna, Rajesh Bhatt, Kevin Gimpel, and Mohit Iyyer. 2024{\natexlab{b}}.
\newblock \href {https://doi.org/10.18653/v1/2024.findings-naacl.49} {{GEE}! grammar error explanation with large language models}.
\newblock In \emph{Findings of the Association for Computational Linguistics: NAACL 2024}, pages 754--781, Mexico City, Mexico. Association for Computational Linguistics.

\bibitem[{Sonkar et~al.(2024)Sonkar, Ni, Chaudhary, and Baraniuk}]{sonkar-etal-2024-pedagogical}
Shashank Sonkar, Kangqi Ni, Sapana Chaudhary, and Richard Baraniuk. 2024.
\newblock \href {https://doi.org/10.18653/v1/2024.findings-emnlp.797} {Pedagogical alignment of large language models}.
\newblock In \emph{Findings of the Association for Computational Linguistics: EMNLP 2024}, pages 13641--13650, Miami, Florida, USA. Association for Computational Linguistics.

\bibitem[{Sonlu et~al.(2024)Sonlu, Bendiksen, Durupinar, and G{\"u}d{\"u}kbay}]{sonlu2024effects}
Sinan Sonlu, Bennie Bendiksen, Funda Durupinar, and U{\u{g}}ur G{\"u}d{\"u}kbay. 2024.
\newblock The effects of embodiment and personality expression on learning in llm-based educational agents.
\newblock \emph{arXiv preprint arXiv:2407.10993}.

\bibitem[{Stahl et~al.(2024{\natexlab{a}})Stahl, Biermann, Nehring, and Wachsmuth}]{stahl-etal-2024-exploring}
Maja Stahl, Leon Biermann, Andreas Nehring, and Henning Wachsmuth. 2024{\natexlab{a}}.
\newblock \href {https://aclanthology.org/2024.bea-1.23/} {Exploring {LLM} prompting strategies for joint essay scoring and feedback generation}.
\newblock In \emph{Proceedings of the 19th Workshop on Innovative Use of NLP for Building Educational Applications (BEA 2024)}, pages 283--298, Mexico City, Mexico. Association for Computational Linguistics.

\bibitem[{Stahl et~al.(2024{\natexlab{b}})Stahl, Biermann, Nehring, and Wachsmuth}]{stahl2024exploring}
Maja Stahl, Leon Biermann, Andreas Nehring, and Henning Wachsmuth. 2024{\natexlab{b}}.
\newblock Exploring llm prompting strategies for joint essay scoring and feedback generation.
\newblock \emph{arXiv preprint arXiv:2404.15845}.

\bibitem[{Stamper et~al.(2024)Stamper, Xiao, and Hou}]{stamper2024enhancing}
John Stamper, Ruiwei Xiao, and Xinying Hou. 2024.
\newblock Enhancing llm-based feedback: Insights from intelligent tutoring systems and the learning sciences.
\newblock In \emph{International Conference on Artificial Intelligence in Education}, pages 32--43. Springer.

\bibitem[{Steinberg and Sciarini(2013)}]{steinberg2013introduction}
Danny~D Steinberg and Natalia~V Sciarini. 2013.
\newblock \emph{An introduction to psycholinguistics}.
\newblock Routledge.

\bibitem[{Syamkumar et~al.(2024)Syamkumar, Tseng, Barron, Yang, Karumbaiah, Uppal, and Hu}]{syamkumar2024improving}
Anand Syamkumar, Nora Tseng, Kaycie Barron, Shanglin Yang, Shamya Karumbaiah, Rheeya Uppal, and Junjie Hu. 2024.
\newblock Improving bilingual capabilities of language models to support diverse linguistic practices in education.
\newblock \emph{arXiv preprint arXiv:2411.04308}.

\bibitem[{Tan et~al.(2024)Tan, Yu, Long, Ma, Murray, Silverman, Yeatman, and Frank}]{tan2024devbench}
Alvin Tan, Chunhua Yu, Bria Long, Wanjing Ma, Tonya Murray, Rebecca Silverman, Jason Yeatman, and Michael~C Frank. 2024.
\newblock Devbench: A multimodal developmental benchmark for language learning.
\newblock \emph{Advances in Neural Information Processing Systems}, 37:77445--77467.

\bibitem[{Taneja(1995)}]{taneja1995educational}
Vidya~Ratna Taneja. 1995.
\newblock \emph{Educational thought and practice}.
\newblock Sterling Publishers Pvt. Ltd.

\bibitem[{Tang et~al.(2024)Tang, Chang, and Yang}]{tang2024pdfchatannotator}
Yi~Tang, Chia-Ming Chang, and Xi~Yang. 2024.
\newblock Pdfchatannotator: A human-llm collaborative multi-modal data annotation tool for pdf-format catalogs.
\newblock In \emph{Proceedings of the 29th International Conference on Intelligent User Interfaces}, pages 419--430.

\bibitem[{Toma(1977)}]{toma1977systran}
Peter Toma. 1977.
\newblock Systran as a multilingual machine translation system.
\newblock In \emph{Proceedings of the Third European Congress on Information Systems and Networks, overcoming the language barrier}, pages 569--581.

\bibitem[{Tong et~al.(2024)Tong, Li, Wang, Wang, Teng, and Shang}]{tong2024can}
Yongqi Tong, Dawei Li, Sizhe Wang, Yujia Wang, Fei Teng, and Jingbo Shang. 2024.
\newblock Can llms learn from previous mistakes? investigating llms' errors to boost for reasoning.
\newblock \emph{arXiv preprint arXiv:2403.20046}.

\bibitem[{T{\"o}rnberg(2024)}]{tornberg2024best}
Petter T{\"o}rnberg. 2024.
\newblock Best practices for text annotation with large language models.
\newblock \emph{arXiv preprint arXiv:2402.05129}.

\bibitem[{Vaswani(2017)}]{vaswani2017attention}
A~Vaswani. 2017.
\newblock Attention is all you need.
\newblock \emph{Advances in Neural Information Processing Systems}.

\bibitem[{Vesselinov and Grego(2012)}]{vesselinov2012duolingo}
Roumen Vesselinov and John Grego. 2012.
\newblock Duolingo effectiveness study.
\newblock \emph{City University of New York, USA}, 28(1-25).

\bibitem[{Wang et~al.(2024{\natexlab{a}})Wang, Zhao, Van~Kleek, and Shadbolt}]{wang2024challenges}
Ge~Wang, Jun Zhao, Max Van~Kleek, and Nigel Shadbolt. 2024{\natexlab{a}}.
\newblock Challenges and opportunities in translating ethical ai principles into practice for children.
\newblock \emph{Nature Machine Intelligence}, 6(3):265--270.

\bibitem[{Wang et~al.(2024{\natexlab{b}})Wang, Macina, Daheim, Pal~Chowdhury, and Sachan}]{wang-etal-2024-book2dial}
Junling Wang, Jakub Macina, Nico Daheim, Sankalan Pal~Chowdhury, and Mrinmaya Sachan. 2024{\natexlab{b}}.
\newblock \href {https://doi.org/10.18653/v1/2024.findings-acl.578} {{B}ook2{D}ial: Generating teacher student interactions from textbooks for cost-effective development of educational chatbots}.
\newblock In \emph{Findings of the Association for Computational Linguistics: ACL 2024}, pages 9707--9731, Bangkok, Thailand. Association for Computational Linguistics.

\bibitem[{Wang et~al.(2024{\natexlab{c}})Wang, Zhu, Ren, Liu, Li, Zhang, Zhang, Wu, Zhan, Liu et~al.}]{wang2024survey}
Ke~Wang, Jiahui Zhu, Minjie Ren, Zeming Liu, Shiwei Li, Zongye Zhang, Chenkai Zhang, Xiaoyu Wu, Qiqi Zhan, Qingjie Liu, and 1 others. 2024{\natexlab{c}}.
\newblock A survey on data synthesis and augmentation for large language models.
\newblock \emph{arXiv preprint arXiv:2410.12896}.

\bibitem[{Wang et~al.(2024{\natexlab{d}})Wang, Zhu, Liu, Zheng, Chen, and Li}]{wang2024knowledge}
Song Wang, Yaochen Zhu, Haochen Liu, Zaiyi Zheng, Chen Chen, and Jundong Li. 2024{\natexlab{d}}.
\newblock Knowledge editing for large language models: A survey.
\newblock \emph{ACM Computing Surveys}, 57(3):1--37.

\bibitem[{Wang et~al.(2025)Wang, Zhan, Lian, Hu, Yuan, Zhang, Xie, and Xiong}]{wang2025llm}
Tianfu Wang, Yi~Zhan, Jianxun Lian, Zhengyu Hu, Nicholas~Jing Yuan, Qi~Zhang, Xing Xie, and Hui Xiong. 2025.
\newblock Llm-powered multi-agent framework for goal-oriented learning in intelligent tutoring system.
\newblock \emph{arXiv preprint arXiv:2501.15749}.

\bibitem[{Wang et~al.(2024{\natexlab{e}})Wang, Kim, Rahman, Mitra, and Miao}]{wang2024human}
Xinru Wang, Hannah Kim, Sajjadur Rahman, Kushan Mitra, and Zhengjie Miao. 2024{\natexlab{e}}.
\newblock Human-llm collaborative annotation through effective verification of llm labels.
\newblock In \emph{Proceedings of the CHI Conference on Human Factors in Computing Systems}, pages 1--21.

\bibitem[{Watzke(2003)}]{watzke2003lasting}
John~L Watzke. 2003.
\newblock \emph{Lasting change in foreign language education: A historical case for change in national policy}.
\newblock Bloomsbury Publishing USA.

\bibitem[{Williams et~al.(2004)Williams, Burden, Poulet, and Maun}]{williams2004learners}
Marion Williams, Robert Burden, G{\'e}rard Poulet, and Ian Maun. 2004.
\newblock Learners' perceptions of their successes and failures in foreign language learning.
\newblock \emph{Language Learning Journal}, 30(1):19--29.

\bibitem[{Willis(2021)}]{willis2021framework}
Jane Willis. 2021.
\newblock \emph{A framework for task-based learning}.
\newblock Intrinsic Books Ltd.

\bibitem[{Wu et~al.(2022)Wu, Xiao, Sun, Zhang, Ma, and He}]{wu2022survey}
Xingjiao Wu, Luwei Xiao, Yixuan Sun, Junhang Zhang, Tianlong Ma, and Liang He. 2022.
\newblock A survey of human-in-the-loop for machine learning.
\newblock \emph{Future Generation Computer Systems}, 135:364--381.

\bibitem[{Xiao et~al.(2024)Xiao, Ma, Song, Xu, Zhang, Wang, and Fu}]{xiao2024humanaicollaborativeessayscoring}
Changrong Xiao, Wenxing Ma, Qingping Song, Sean~Xin Xu, Kunpeng Zhang, Yufang Wang, and Qi~Fu. 2024.
\newblock \href {https://arxiv.org/abs/2401.06431} {Human-ai collaborative essay scoring: A dual-process framework with llms}.
\newblock \emph{Preprint}, arXiv:2401.06431.

\bibitem[{Xiao et~al.(2023)Xiao, Xu, Zhang, Wang, and Xia}]{xiao-etal-2023-evaluating}
Changrong Xiao, Sean~Xin Xu, Kunpeng Zhang, Yufang Wang, and Lei Xia. 2023.
\newblock \href {https://doi.org/10.18653/v1/2023.bea-1.52} {Evaluating reading comprehension exercises generated by {LLM}s: A showcase of {C}hat{GPT} in education applications}.
\newblock In \emph{Proceedings of the 18th Workshop on Innovative Use of NLP for Building Educational Applications (BEA 2023)}, pages 610--625, Toronto, Canada. Association for Computational Linguistics.

\bibitem[{Xu et~al.(2023)Xu, Huang, Liu, Shen, Liu, Chen, Wu, and Wang}]{xu2023learning}
Bihan Xu, Zhenya Huang, Jiayu Liu, Shuanghong Shen, Qi~Liu, Enhong Chen, Jinze Wu, and Shijin Wang. 2023.
\newblock Learning behavior-oriented knowledge tracing.
\newblock In \emph{Proceedings of the 29th ACM SIGKDD conference on knowledge discovery and data mining}, pages 2789--2800.

\bibitem[{Xu et~al.(2024)Xu, Zhang, and Qin}]{xu2024eduagent}
Songlin Xu, Xinyu Zhang, and Lianhui Qin. 2024.
\newblock Eduagent: Generative student agents in learning.
\newblock \emph{arXiv preprint arXiv:2404.07963}.

\bibitem[{Yao et~al.(2024)Yao, Parashar, Zhou, Jang, Ouyang, Yang, and Yu}]{yao2024mcqg}
Zonghai Yao, Aditya Parashar, Huixue Zhou, Won~Seok Jang, Feiyun Ouyang, Zhichao Yang, and Hong Yu. 2024.
\newblock Mcqg-srefine: Multiple choice question generation and evaluation with iterative self-critique, correction, and comparison feedback.
\newblock \emph{arXiv preprint arXiv:2410.13191}.

\bibitem[{Ye et~al.(2023)Ye, Li, Li, and Zheng}]{ye-etal-2023-mixedit}
Jingheng Ye, Yinghui Li, Yangning Li, and Hai-Tao Zheng. 2023.
\newblock \href {https://doi.org/10.18653/v1/2023.findings-emnlp.681} {{M}ix{E}dit: Revisiting data augmentation and beyond for grammatical error correction}.
\newblock In \emph{Findings of the Association for Computational Linguistics: EMNLP 2023}, pages 10161--10175, Singapore. Association for Computational Linguistics.

\bibitem[{Ye et~al.(2024)Ye, Qin, Li, Cheng, Qin, Zheng, Xing, Xu, Cheng, and Wei}]{ye2024excgec}
Jingheng Ye, Shang Qin, Yinghui Li, Xuxin Cheng, Libo Qin, Hai-Tao Zheng, Peng Xing, Zishan Xu, Guo Cheng, and Zhao Wei. 2024.
\newblock Excgec: A benchmark of edit-wise explainable chinese grammatical error correction.
\newblock \emph{arXiv preprint arXiv:2407.00924}.

\bibitem[{Yu et~al.(2019)Yu, Si, Hu, and Zhang}]{yu2019review}
Yong Yu, Xiaosheng Si, Changhua Hu, and Jianxun Zhang. 2019.
\newblock A review of recurrent neural networks: Lstm cells and network architectures.
\newblock \emph{Neural computation}, 31(7):1235--1270.

\bibitem[{Yue et~al.(2024)Yue, Mifdal, Zhang, Suh, and Yao}]{yue2024mathvc}
Murong Yue, Wijdane Mifdal, Yixuan Zhang, Jennifer Suh, and Ziyu Yao. 2024.
\newblock Mathvc: An llm-simulated multi-character virtual classroom for mathematics education.
\newblock \emph{arXiv preprint arXiv:2404.06711}.

\bibitem[{Zeng et~al.(2023)Zeng, Zhao, He, Geng, Wang, Wu, and Xu}]{zeng-etal-2023-seen}
Weihao Zeng, Lulu Zhao, Keqing He, Ruotong Geng, Jingang Wang, Wei Wu, and Weiran Xu. 2023.
\newblock \href {https://doi.org/10.18653/v1/2023.acl-long.793} {Seen to unseen: Exploring compositional generalization of multi-attribute controllable dialogue generation}.
\newblock In \emph{Proceedings of the 61st Annual Meeting of the Association for Computational Linguistics (Volume 1: Long Papers)}, pages 14179--14196, Toronto, Canada. Association for Computational Linguistics.

\bibitem[{Zeng et~al.(2024)Zeng, Yu, Gao, Meng, Goyal, and Chen}]{zeng2024evaluating}
Zhiyuan Zeng, Jiatong Yu, Tianyu Gao, Yu~Meng, Tanya Goyal, and Danqi Chen. 2024.
\newblock \href {https://openreview.net/forum?id=tr0KidwPLc} {Evaluating large language models at evaluating instruction following}.
\newblock In \emph{The Twelfth International Conference on Learning Representations}.

\bibitem[{Zha et~al.(2023)Zha, Bhat, Lai, Yang, Jiang, Zhong, and Hu}]{zha2023data}
Daochen Zha, Zaid~Pervaiz Bhat, Kwei-Herng Lai, Fan Yang, Zhimeng Jiang, Shaochen Zhong, and Xia Hu. 2023.
\newblock Data-centric artificial intelligence: A survey.
\newblock \emph{ACM Computing Surveys}.

\bibitem[{Zhan et~al.(2024)Zhan, Guo, Li, Hou, Liang, Gao, Luo, and Liu}]{zhan2024knowledge}
Bojun Zhan, Teng Guo, Xueyi Li, Mingliang Hou, Qianru Liang, Boyu Gao, Weiqi Luo, and Zitao Liu. 2024.
\newblock Knowledge tracing as language processing: A large-scale autoregressive paradigm.
\newblock In \emph{International Conference on Artificial Intelligence in Education}, pages 177--191. Springer.

\bibitem[{Zhang et~al.(2023)Zhang, Song, Li, Zhou, and Song}]{zhang2023survey}
Hanqing Zhang, Haolin Song, Shaoyu Li, Ming Zhou, and Dawei Song. 2023.
\newblock A survey of controllable text generation using transformer-based pre-trained language models.
\newblock \emph{ACM Computing Surveys}, 56(3):1--37.

\bibitem[{Zhang et~al.(2024{\natexlab{a}})Zhang, Yao, Tian, Wang, Deng, Wang, Xi, Mao, Zhang, Ni et~al.}]{zhang2024comprehensive}
Ningyu Zhang, Yunzhi Yao, Bozhong Tian, Peng Wang, Shumin Deng, Mengru Wang, Zekun Xi, Shengyu Mao, Jintian Zhang, Yuansheng Ni, and 1 others. 2024{\natexlab{a}}.
\newblock A comprehensive study of knowledge editing for large language models.
\newblock \emph{arXiv preprint arXiv:2401.01286}.

\bibitem[{Zhang et~al.(2021)Zhang, Guo, Chen, Fan, and Cheng}]{zhang2021review}
Ruqing Zhang, Jiafeng Guo, Lu~Chen, Yixing Fan, and Xueqi Cheng. 2021.
\newblock A review on question generation from natural language text.
\newblock \emph{ACM Transactions on Information Systems (TOIS)}, 40(1):1--43.

\bibitem[{Zhang et~al.(2024{\natexlab{b}})Zhang, Zhang-Li, Yu, Gong, Zhou, Liu, Hou, and Li}]{zhang2024simulating}
Zheyuan Zhang, Daniel Zhang-Li, Jifan Yu, Linlu Gong, Jinchang Zhou, Zhiyuan Liu, Lei Hou, and Juanzi Li. 2024{\natexlab{b}}.
\newblock Simulating classroom education with llm-empowered agents.
\newblock \emph{arXiv preprint arXiv:2406.19226}.

\bibitem[{Zhang-Li et~al.(2024)Zhang-Li, Zhang, Yu, Yin, Tu, Gong, Wang, Liu, Liu, Hou et~al.}]{zhang2024awaking}
Daniel Zhang-Li, Zheyuan Zhang, Jifan Yu, Joy Lim~Jia Yin, Shangqing Tu, Linlu Gong, Haohua Wang, Zhiyuan Liu, Huiqin Liu, Lei Hou, and 1 others. 2024.
\newblock Awaking the slides: A tuning-free and knowledge-regulated ai tutoring system via language model coordination.
\newblock \emph{arXiv preprint arXiv:2409.07372}.

\bibitem[{Zheng et~al.(2024)Zheng, Li, Huang, Liang, Guo, Hou, Gao, Tian, Liu, and Luo}]{zheng2024automatic}
Ying Zheng, Xueyi Li, Yaying Huang, Qianru Liang, Teng Guo, Mingliang Hou, Boyu Gao, Mi~Tian, Zitao Liu, and Weiqi Luo. 2024.
\newblock Automatic lesson plan generation via large language models with self-critique prompting.
\newblock In \emph{International Conference on Artificial Intelligence in Education}, pages 163--178. Springer.

\bibitem[{Zhong et~al.(2024)Zhong, Guo, Gao, Ye, and Wang}]{zhong2024memorybank}
Wanjun Zhong, Lianghong Guo, Qiqi Gao, He~Ye, and Yanlin Wang. 2024.
\newblock Memorybank: Enhancing large language models with long-term memory.
\newblock In \emph{Proceedings of the AAAI Conference on Artificial Intelligence}, volume~38, pages 19724--19731.

\bibitem[{Zhou et~al.(2024)Zhou, Chen, and Yu}]{zhou-etal-2024-llm}
Ruiyang Zhou, Lu~Chen, and Kai Yu. 2024.
\newblock \href {https://aclanthology.org/2024.lrec-main.816/} {Is {LLM} a reliable reviewer? a comprehensive evaluation of {LLM} on automatic paper reviewing tasks}.
\newblock In \emph{Proceedings of the 2024 Joint International Conference on Computational Linguistics, Language Resources and Evaluation (LREC-COLING 2024)}, pages 9340--9351, Torino, Italia. ELRA and ICCL.

\bibitem[{Zhou et~al.(2025)Zhou, Zhang, Jiang, Gao, Liu, and Jiang}]{zhou2025study}
Yizhou Zhou, Mengqiao Zhang, Yuan-Hao Jiang, Xinyu Gao, Naijie Liu, and Bo~Jiang. 2025.
\newblock A study on educational data analysis and personalized feedback report generation based on tags and chatgpt.
\newblock \emph{arXiv preprint arXiv:2501.06819}.

\bibitem[{Zhuang et~al.(2024)Zhuang, Wu, Shen, Yu, Yi, Chen, Hu, Chen, Ren, Zhang, Song, Liu, and Lan}]{zhuang-etal-2024-toree}
Xinlin Zhuang, Hongyi Wu, Xinshu Shen, Peimin Yu, Gaowei Yi, Xinhao Chen, Tu~Hu, Yang Chen, Yupei Ren, Yadong Zhang, Youqi Song, Binxuan Liu, and Man Lan. 2024.
\newblock \href {https://doi.org/10.18653/v1/2024.findings-acl.342} {{TOREE}: Evaluating topic relevance of student essays for {C}hinese primary and middle school education}.
\newblock In \emph{Findings of the Association for Computational Linguistics: ACL 2024}, pages 5749--5765, Bangkok, Thailand. Association for Computational Linguistics.

\end{thebibliography}

\appendix

\section{Literature Review}
\label{app:review}
We provide an overview of LLM-centric research of English Education presented in Figure~\ref{fig:review}.

\tikzstyle{edge}=[-latex',draw=black!90,shorten <=1pt,shorten >=1pt]
\tikzstyle{redge}=[latex'-,draw=black!90,shorten <=1pt,shorten >=1pt]
\tikzstyle{dedge}=[latex'-latex',draw=black!90,shorten <=1pt,shorten >=1pt]

\tikzstyle{block}=[draw, text width=5em,align=center,shape=rectangle, rounded corners, , align=center]
\tikzstyle{nobox}=[align=center]
\definecolor{emb}{RGB}{209,228,252}
\definecolor{hidden-blue}{RGB}{194,232,247}
\definecolor{hidden-orange}{RGB}{224,224,224}
\definecolor{hidden-yellow}{RGB}{242,244,193}
\definecolor{output-purple}{RGB}{219,203,231}
\definecolor{output-green}{RGB}{204,231,207}
\definecolor{output-blue}{RGB}{44,169,225}

\definecolor{output-black}{RGB}{0,0,0}
\definecolor{output-white}{RGB}{255,255,255}
\definecolor{myblue}{RGB}{137,195,235}
\definecolor{hiddendraw}{RGB}{137,195,235}

\tikzstyle{leaf}=[draw=hiddendraw,
    rounded corners, minimum height=1em,
    fill=myblue!40,text opacity=1, 
    fill opacity=.5,  text=black,align=left,font=\scriptsize,
    inner xsep=3pt,
    inner ysep=1pt,
]
\tikzstyle{middle}=[draw=hiddendraw,
    rounded corners, minimum height=1.5em,
    fill=output-white!40,text opacity=1, 
    fill opacity=.5, text=black, align=center, font=\small,
    inner xsep=7pt,
    inner ysep=1pt,
]

\begin{figure*}[htbp!]
\centering
\begin{forest}
  for tree={
      forked edges,
      grow=east,
      reversed=true,
      anchor=base west,
      parent anchor=east,
      child anchor=west,
      base=middle,
      font=\scriptsize,
      rectangle,
      line width=0.8pt,
      draw=output-black,
      rounded corners,align=left,
      minimum width=2em, s sep=6pt, l sep=8pt,
  },
  where level=1{text width=0.2\linewidth}{},
  where level=2{text width=0.2\linewidth,font=\scriptsize}{},
  where level=3{font=\scriptsize}{},
  where level=4{font=\scriptsize}{},
  where level=5{font=\scriptsize}{},
  [LLMs for English Education, middle,rotate=90,anchor=north,edge=output-black
      [LLM as Data Enhancer\\(Section \ref{sec:enhancer}),middle,anchor=west,edge=output-black, text width=0.20\linewidth
        [Data Creation, middle, text width=0.14\linewidth, edge=output-black
            [{
            PFQS~\cite{li-zhang-2024-planning},
            MCQG-SRefine~\cite{yao2024mcqg},\\
            \cite{lee2024few},
            BF-TC~\cite{liu2024personality},
            MathVC~\cite{yue2024mathvc},\\
            EduAgent~\cite{xu2024eduagent},
            Generative Students~\cite{lu2024generative}
            }, leaf, text width=0.48\linewidth, edge=output-black]
        ]
        [Data Reformation, middle, text width=0.14\linewidth, edge=output-black
            [{
            Book2Dial~\cite{wang-etal-2024-book2dial}, Slide2Lecture~\cite{zhang2024awaking},\\
            WikiDomains~\cite{asthana-etal-2024-evaluating}, \cite{freyer2024easy},\\
            \cite{day2025evaluating},
            Anthropological Prompting~\cite{alkhamissi-etal-2024-investigating},\\
            \cite{liu2024culturally}
            }, leaf, text width=0.48\linewidth, edge=output-black]
        ]
        [Data Annotation, middle, text width=0.14\linewidth, edge=output-black
            [{
            MEGAnno+~\cite{kim-etal-2024-meganno},
            EDEN~\cite{li-etal-2024-eden},\\
            PDFChatAnnotator~\cite{tang2024pdfchatannotator},
            Coannotating~\cite{li-etal-2023-coannotating}
            }, leaf, text width=0.48\linewidth, edge=output-black]
        ]
      ]
      [LLM as Task Predictor\\(Section \ref{sec:predictor}),middle,anchor=west,edge=output-black, text width=0.20\linewidth
        [Discriminative, middle, text width=0.14\linewidth, edge=output-black
            [{
            TOREE~\cite{zhuang-etal-2024-toree},
            LLM-KT~\cite{zhan2024knowledge},\\
            CLST~\cite{jung2024clst},
            Diallogue-KT~\cite{scarlatos2024exploring},\\
            \cite{neshaei2024towards},
            \cite{mizumoto2023exploring},
            \cite{song-etal-2025-unified},\\
            \cite{sessler2024can},
            \cite{syamkumar2024improving}
            }, leaf, text width=0.48\linewidth, edge=output-black]
        ]
        [Generative, middle, text width=0.14\linewidth, edge=output-black
            [{
            RECIPE~\cite{han2023recipe},
            EXPECT~\cite{fei-etal-2023-enhancing},\\
            GEE~\cite{song-etal-2024-gee},
            EXCGEC~\cite{ye2024excgec},\\
            FELT~\cite{borges-etal-2024-teach},
            LLM-as-a-tutor~\cite{han-etal-2024-llm},\\
            SocraticLM~\cite{liusocraticlm},
            \cite{stamper2024enhancing},\\
            \cite{favero2024enhancing},
            \cite{xiao2024humanaicollaborativeessayscoring},
            \cite{scarlatos2025training}
            }, leaf, text width=0.48\linewidth, edge=output-black]
        ]
        [Mixed, middle, text width=0.14\linewidth, edge=output-black
            [{
            FABRIC~\cite{han2023fabric},
            eRevise+RF~\cite{liu2025erevise+},\\
            ReaLMistake~\cite{kamoi2024evaluating},
            \cite{stahl2024exploring},
            \cite{lu2023error}
            }, leaf, text width=0.48\linewidth, edge=output-black]
        ]
      ]
      [LLM-empowered Agent\\(Section \ref{sec:agent}),middle,anchor=west,edge=output-black, text width=0.20\linewidth
        [Abilities, middle, text width=0.14\linewidth, edge=output-black
            [{
            FOKE~\cite{hu2024foke},
            KnowEdit~\cite{zhang2024comprehensive},\\
            PedCoT~\cite{jiang2024llms},
            LessonPlanner~\cite{fan2024lessonplanner},\\
            MemoryBank~\cite{zhong2024memorybank},
            vNMF~\cite{jiang2024ai},\\
            ChatTutor~\cite{chen2024empowering},
            LHP~\cite{sonkar-etal-2024-pedagogical},\\
            \cite{razafinirina2024pedagogical},
            \cite{zhang2024comprehensive},\\
            \cite{hu2024teaching},
            \cite{zheng2024automatic},
            \cite{shi2025educationq}
            }, leaf, text width=0.48\linewidth, edge=output-black]
        ]
        [Applications, middle, text width=0.14\linewidth, edge=output-black
            [{
            SimClass~\cite{zhang2024simulating}, MathVC~\cite{yue2024mathvc},\\
            BIPED~\cite{kwon-etal-2024-biped},
            LLM-as-a-tutor~\cite{han-etal-2024-llm},\\
            KORLINGS~\cite{lee2024developing},
            \cite{liu2024personality}
            }, leaf, text width=0.48\linewidth, edge=output-black]
        ]
      ]
]
\end{forest}
\caption{An overview of LLM-centric research of FLE.}
\label{fig:review}
\end{figure*}

\section{Future Directions}
\label{app:future_direction}
\paragraph{Establishing Robust Evaluation Frameworks.}
A significant challenge in leveraging LLMs for English Education is the current lack of widely accepted and easily implementable evaluation frameworks to assess the quality of LLM-based teaching interactions and outcomes. Existing metrics often focus on linguistic correctness or task completion~\cite{tan2024devbench,macina2025mathtutorbench} rather than pedagogical efficacy or impact on learning~\cite{chiang-etal-2024-large}. Future work should prioritize the development of standardized evaluation methodologies, including comprehensive benchmarks and nuanced metrics that capture both the accuracy of linguistic information and the pedagogical value of LLM interventions. This will be essential for comparing different systems and guiding iterative improvements.

\paragraph{Integrating with Modern Standardized Educational Frameworks.}
English language learning is often governed by established standards and frameworks, such as the Common European Framework of Reference for Languages (CEFR)\footnote{\url{https://www.coe.int/en/web/portal/home}} or Common Core State Standards (CCSS)\footnote{\url{https://corestandards.org/}}. For LLM-based tools to be truly effective and gain acceptance, their outputs and interaction patterns should align with these existing frameworks. Future technical development should focus on enabling LLMs to reference, interpret, and operate consistently within these standards~\cite{nicholls2024write,imperial-etal-2024-standardize}. This includes generating proficiency-level-appropriate content, providing feedback that corresponds to specific framework descriptors, and assisting learners in achieving standardized learning objectives, thereby enhancing usability, conformity, and trustworthiness among educators and learners.

\paragraph{Fostering Human-AI Collaboration in Pedagogy.}
While LLMs offer transformative potential, it is unlikely they will completely replace human teachers in English Education in the foreseeable future. Instead, the most promising path involves developing sophisticated human-AI collaborative educational technologies~\cite{kim2025augmented}. Future research should explore how LLMs can best function as assistive tools that augment, rather than supplant, the capabilities of human educators~\cite{shojaei2025ai}. This includes designing intuitive interfaces for teachers to guide, customize, and oversee LLM-driven activities, investigating teachers' perspectives on integrating LLMs into their practice, and defining technical benchmarks for when an LLM possesses sufficient acquired skills to reliably assist teachers. The focus must be on a synergistic model where LLMs handle scalable tasks while human teachers provide the crucial elements of empathy, nuanced understanding, and holistic student development.

\section{Alternative Views}\label{sec:alternative_views}
While this paper supports the use of LLMs in English Education, it is essential to consider alternative perspectives. Below, we discuss two key opposing views and provide counterarguments.

\subsection{Task-Specific or Language-Specific Models as Better Alternatives}
Some argue that specialized or language-specific models, including classical ML systems with carefully engineered features, can outperform general-purpose LLMs in narrowly defined tasks (e.g., phonetics or grammar drills~\cite{fang2023chatgpt}). By focusing on limited objectives, such models avoid the computational overhead and potential inaccuracies of LLMs, which aim to handle a broader range of inputs and contexts~\cite{shen2024language}.

\paragraph{Counterargument.}
While specialized models may excel in isolated tasks, they lack the flexibility required for comprehensive English Education, which involves cultural nuances, conversations, and evolving learner needs. In contrast, LLMs can be fine-tuned for specific goals while still offering broader linguistic competence~\cite{song2024multilingual}. Additionally, relying on multiple specialized models can be resource-intensive, whereas a well-configured LLM provides a unified framework that balances specialization and scalability.

\subsection{Concerns About Over-Reliance on LLMs}
Critics warn that over-reliance on LLMs may lead to problems such as generating misleading outputs~\cite{nahar2024fakes}, reducing human interaction, and over-standardizing teaching methods. These issues could undermine the interpersonal and motivational aspects of language learning.

\paragraph{Counterargument.}
These risks highlight the need for balanced integration rather than the replacement of human tutors. LLMs can complement educators by automating repetitive tasks, allowing teachers to focus on individualized support and motivation. Advances in AI safety, such as feedback loops~\cite{tong2024can} and human-in-the-loop systems~\cite{wu2022survey}, can help minimize inaccuracies~\cite{ho2024mitigating}. Additionally, the fine-tuning capabilities of LLMs ensure adaptability, supporting diverse and inclusive learning experiences~\cite{lee2024human}.


\end{document}